\PassOptionsToPackage{section}{placeins}
\documentclass[]{ailab}
\usepackage{tabularx}
\usepackage{url}
\usepackage{adjustbox}
\usepackage{multirow}
\usepackage{pifont}
\usepackage{comment}
\usepackage{amsmath,amssymb}
\usepackage{amsthm}
\usepackage{mathtools}
\usepackage{colortbl}
\usepackage{color}
\usepackage{booktabs}
\setcitestyle{square,comma,numbers,sort&compress}
\usepackage{graphicx}
\usepackage{wrapfig}
\usepackage{xspace}
\usepackage{nicefrac}
\usepackage{enumitem}
\usepackage{float}
\usepackage[section]{placeins}
\usepackage{CJKutf8}
\usepackage{tikz}

\usepackage{xcolor}
\usepackage{colortbl}

\definecolor{bestbg}{RGB}{225,238,255}
\definecolor{worstbg}{RGB}{255,232,218}

\newcommand{\bestcell}[1]{\cellcolor{bestbg}\textbf{#1}}
\newcommand{\worstcell}[1]{\cellcolor{worstbg}\textbf{#1}}

\definecolor{darkblue}{rgb}{0,0,0.5}
\hypersetup{
  colorlinks=true,
  citecolor=darkblue,
  linkcolor=darkblue,
  urlcolor=darkblue
}

\crefname{appendix}{Appendix}{appendices}
\Crefname{appendix}{Appendix}{Appendices}

\definecolor{lightgray}{gray}{0.95}
\definecolor{deepgray}{RGB}{119,136,153}
\tcbset{
  aibox/.style={
    enhanced jigsaw,
    breakable,
    colback=metabg!5,
    colframe=deepgray,
    colbacktitle=metabg,
    coltitle=white,
    boxrule=0.45pt,
    arc=1.5pt,
    left=6pt,
    right=6pt,
    top=5pt,
    bottom=5pt,
    before skip=5pt,
    after skip=7pt,
    fonttitle=\sffamily\bfseries\small,
    fontupper=\sffamily\fontsize{8pt}{10pt}\selectfont
  }
}
\newtcolorbox{AIbox}[2][]{aibox,title=#2,#1}

\definecolor{patternnt}{RGB}{70,91,112}
\definecolor{patternt}{RGB}{93,73,112}
\newcommand{\smalltag}[2]{%
  \tikz[baseline=(X.base)]{
    \node[
      rounded corners=2.5pt,
      fill=#1!10,
      draw=#1!28,
      text=#1!70!black,
      line width=0.3pt,
      inner xsep=3.2pt,
      inner ysep=0.8pt,
      font=\sffamily\bfseries\tiny
    ] (X) {#2};
  }%
}
\newcommand{\NTicon}{\smalltag{patternnt}{Non-thk}}
\newcommand{\Ticon}{\smalltag{patternt}{Thk}}

\renewcommand{\paragraph}[1]{\vspace{1.25mm}\noindent\textbf{#1}}
\title{Beyond Correctness: Benchmarking and Aligning Response Behaviors in Hybrid-Thinking MLLMs}

\author[1,2, 5,*]{Xinming Wang}
\author[2,\dagger]{Weinong Wang}
\author[2]{Hongming Yang}
\author[3]{Yansong Lin}
\author[2]{Zheng Ruan}
\author[2,4]{Shangpin Peng}
\author[2]{Qiming Peng}
\author[2]{Nan Qiao}
\author[2]{Fengyuan Lu}
\author[2]{Guoqing Ma}
\author[2]{Marito Li}
\author[2]{Songyang Zhang}
\author[2]{Saiyong Yang}
\author[2]{Han Hu}
\author[2]{Yonglong Tian}
\author[1,\dagger]{Xu-Yao Zhang}

\affiliation[1]{Institute of Automation, Chinese Academy of Sciences\\}
\affiliation[2]{Large Language Model Department, Tencent\\}
\affiliation[3]{University of Electronic Science and Technology of China\\}
\affiliation[4]{Hong Kong University of Science and Technology\\}
\affiliation[5]{Zhongguancun Academy}

\contribution[*]{This work was done during an internship at Tencent.}
\contribution[\dagger]{Corresponding authors.}

\email{wangxinming2024@ia.ac.cn, \{weinongwang,hmingyang\}@tencent.com, xyz@nlpr.ia.ac.cn}

\abstract{
Hybrid-thinking multimodal large language models (MLLMs) allow a single model to alternate between deliberative thinking and latency-efficient non-thinking inference. Although these modes differ in reasoning budget, their delivered responses should satisfy the same user-facing standard. Correctness alone may not characterize this response quality; we therefore evaluate task accuracy and response-pattern failures as complementary outcomes.
We study this gap through \textbf{response-pattern alignment}: whether thinking and non-thinking interfaces preserve acceptable final-response behavior. We introduce \textbf{PatternEval}, a failure-enriched diagnostic benchmark comprising 2,415 multimodal prompts spanning visual perception and grounding, structured image understanding, and multimodal knowledge reasoning. PatternEval tests four recurrent failures: chain-of-thought leakage, response repetition, logical contradiction, and performative reasoning. Response-pattern failures are widespread across models from different providers, with non-thinking inference exhibiting substantially higher failure rates and thereby creating systematic misalignment between thinking and non-thinking interfaces.
Motivated by this diagnosis, we develop \textbf{PatternRM}, a response-level reward model, and \textbf{PatternRL}, which introduces pattern-specific penalties during reinforcement learning. Experiments on Qwen3-VL-4B and Qwen3-VL-8B show that incorporating pattern-specific penalties into reinforcement learning can mitigate cross-mode misalignment while incurring a marginal task performance trade-off. Together, PatternEval and PatternRL provide an evaluation-and-training framework for aligning user-visible response patterns across hybrid-thinking interfaces.
}

\date{Jul 28, 2026}

\begin{document}
\thispagestyle{firstheader}
\maketitle

\section{Introduction}
\label{sec:introduction}

Multimodal large language models (MLLMs) increasingly serve both reasoning-intensive tasks and latency-sensitive interactions. Modern systems therefore expose \emph{hybrid-thinking} interfaces: a deliberative thinking mode allocates additional test-time computation, while a non-thinking mode returns a direct answer under tighter latency and token budgets \citep{deepseek2025r1,qwen2025qwen3,he2025skywork,glm45team2025glm45,xu2025redstar}. This interface makes reasoning effort controllable without maintaining separate models, but it also creates different user-facing behaviors under different effort, which should remain reliable under the same deployment standard.

The trade-off between capability and efficiency does not justify a trade-off in final-response quality. Regardless of the inference route, an answer should remain grounded, coherent, appropriately concise, and free of unintended process narration. Correctness-driven post-training can steer models toward expected answers without directly constraining all of these response properties: generated text may still expose reasoning-like traces, repeat content, retain incompatible statements, or make unsupported factual claims~\citep{holtzman2019curious,welleck2019neural,manakul2023selfcheckgpt,gan2026tnt}. We call a mode-dependent difference in such user-visible behavior \textbf{response-pattern misalignment}. We therefore ask a question complementary to answer correctness: when the inference mode changes, does the model preserve an acceptable and mode-consistent response pattern?

\begin{figure}[H]
    \centering
    \includegraphics[width=\linewidth]{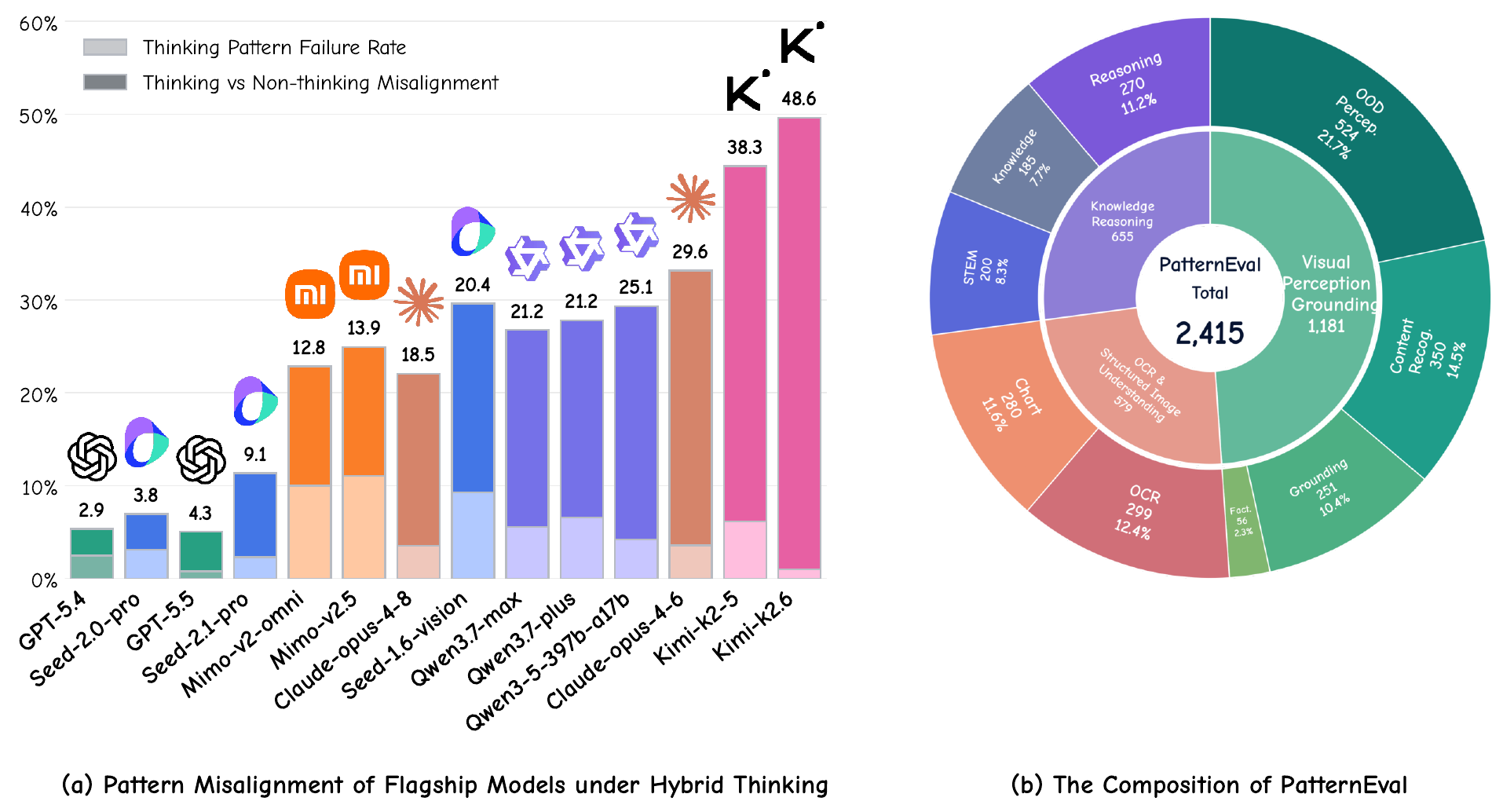}
    \caption{\textbf{Response-pattern misalignment and PatternEval.} (a) Response-pattern misalignment in flagship models under hybrid-thinking configurations. (b) Composition of PatternEval.}

    \label{fig:think-vs-nothink-anyfailure}
\end{figure}

To this end, we introduce \textbf{PatternEval}, a failure-enriched diagnostic benchmark comprising 2,415 multimodal prompts drawn from nine constituent categories. These categories are organized into three broad task families: visual perception and grounding, OCR and structured-image understanding, and multimodal knowledge reasoning. PatternEval deliberately concentrates difficult prompts that expose response-pattern failures, allowing it to stress-test whether models preserve acceptable user-facing behavior under matched thinking and non-thinking interfaces. It captures four recurrent failures: chain-of-thought leakage, response repetition, logical contradiction, and performative reasoning.
Despite substantial advances in overall model capability, these gains do not transfer uniformly across inference interfaces. As shown in \Cref{fig:think-vs-nothink-anyfailure}, frontier models still exhibit pronounced thinking–non-thinking misalignment, with non-thinking Trigger rates consistently higher and gaps reaching 48.64\%. This suggests that strong task performance does not necessarily guarantee stable response behavior across different patterns, exposing an interface-dependent robustness gap that conventional accuracy metrics may overlook.

These outcomes motivate the addition of response-pattern objectives to the post-training process, which does not directly constrain every property of the delivered response~\citep{amodei2016concrete,skalse2022defining,gao2022scaling,gan2026tnt}. We therefore train \textbf{PatternRM} to recognize the four response-pattern failures and incorporate its category-specific penalties into \textbf{PatternRL}. On Qwen3-VL-4B and Qwen3-VL-8B, PatternRL reduces non-thinking Trigger relative to correctness-only BaseRL by 13.08 and 14.35 percentage points, while aggregate accuracy changes by less than one percentage point. Further experiments show that PatternRL can also be integrated into broader task training, consistently mitigating response-pattern failures across diverse settings and increasing the proportion of usable model responses without materially compromising task performance.

We highlight the following contributions.
\begin{itemize}[leftmargin=*]

\item We formulate \textbf{response-pattern alignment} as a requirement complementary to answer correctness and introduce \textbf{PatternEval}, a failure-enriched multimodal challenge set that stress-tests four user-visible response failures under hybrid-thinking interfaces.

\item Through matched-mode evaluation across diverse hybrid-thinking MLLMs, we identify a consistent thinking--non-thinking misalignment: non-thinking mode exhibits higher Trigger rates, with substantial gaps persisting even among frontier models.

\item We translate response-pattern evaluation into an explicit post-training objective through \textbf{PatternRM} and \textbf{PatternRL}. Compared with correctness-only BaseRL, PatternRL substantially reduces response-pattern failures while preserving aggregate accuracy, and further improves response usability when integrated into broader task training.

\end{itemize}
\section{Related Work}
\label{sec:related-work}

\paragraph{Controlling hybrid thinking and reasoning.}
Chain-of-thought prompting and self-consistency showed that eliciting or sampling intermediate reasoning can improve answer accuracy \citep{wei2022chain,kojima2022large,wang2022selfconsistency}. Reasoning-specialized models then made deliberation explicit test-time computation \citep{deepseek2025r1,he2025skywork,xu2025redstar}, while Qwen3 and GLM-4.5 expose thinking and direct-response modes within a single model \citep{qwen2025qwen3,glm45team2025glm45}. Work on controlling this computation follows three routes: learned routing or mode tokens decide \emph{whether} to reason \citep{fang2025thinkless,wang2025thinkornot}; budget-aware control and adaptive long/short policies regulate \emph{how much} reasoning to allocate \citep{wen2025budgetthinker,luo2025autol2s,zhang2025longshortcot}; and data-centric or architecture-level designs seek stronger behavioral separation \citep{wang2025demystifying,wang2026pathlock}. Direct answers can remain competitive under tight budgets, so longer deliberation is not uniformly preferable \citep{ma2025withoutthinking}. However, these methods primarily optimize accuracy, efficiency, or routing rewards rather than the delivered response. Reasoning may still leak into non-thinking outputs, and a policy can satisfy a non-thinking reward while deliberating visibly \citep{gan2026tnt}. Thus, controlling when and how much to reason leaves unresolved whether the selected mode produces a behaviorally appropriate final response.

\paragraph{Response-pattern evaluation beyond correctness.}
This post-routing question connects reasoning control to evaluation beyond correctness. General benchmarks assess robustness, calibration, truthfulness, safety, and instruction following \citep{liang2022helm,lin2021truthfulqa,rottger2024safetyprompts,zhou2023instruction}; multimodal suites add expert reasoning, hallucination, visual consistency, and trustworthiness \citep{fu2023mme,liu2023mmbench,yue2023mmmu,li2023pope,guan2023hallusionbench,wang2023amber,zhang2025multitrustx}. Instruction-following evaluation asks whether outputs satisfy explicit constraints, while studies of text degeneration, repetition, and hallucination expose recurrent response-level failures \citep{holtzman2019curious,welleck2019neural,manakul2023selfcheckgpt}. Preference datasets and reward-model benchmarks extend evaluation to helpfulness, visual faithfulness, safety, and judgment quality \citep{li2024vlfeedback,lambert2024rewardbench,malik2025rewardbench2,yasunaga2025multimodalrewardbench}, but aggregate scores can obscure specific failures. Model-based evaluation scales such judgments, yet introduces length, self-preference, position, and superficial-reflection biases \citep{liu2023geval,zheng2023judging,dubois2024lengthcontrolled,panickssery2024selfpreference,wwn2024sedareval,wang2025judgingbias}. PatternEval instead treats the response pattern itself as the evaluation object, separating leakage, repetition, contradiction, and unsupported performative reasoning under an auditable taxonomy. PatternRM then converts these failure labels into training signals, closing the loop from post-routing diagnosis to policy optimization.

\section{PatternEval}
\label{sec:patterneval}

\subsection{Benchmark Construction and Composition}
\label{sec:patterneval-design}
\label{sec:patterneval-statistics}

\paragraph{Construction Process.}
PatternEval is deliberately constructed as a failure-enriched diagnostic benchmark from a broad candidate pool of image--prompt pairs spanning visual perception and grounding, OCR and structured-image understanding, and multimodal knowledge reasoning. Its construction follows three sequential steps.
First, we generate rollout responses for the candidate prompts and identify recurrent user-visible failures, with particular attention to difficult non-thinking responses. These observations provide empirical anchors for the benchmark.
Second, we expand from the initial cases to related tasks, prompt structures, and response forms, covering broader manifestations of each failure pattern rather than isolated examples. Third, we apply pattern-oriented sampling and constituent-category balancing: categories dominated by deterministic answer-format checks are down-sampled, categories with meaningful response variation are preferentially retained, and high-volume source categories are capped to prevent them from dominating aggregate results. The resulting 2,415 image--prompt pairs concentrate evaluation on conditions that expose the targeted failures while retaining broad multimodal task coverage. PatternEval therefore measures conditional robustness on a selected stress test rather than natural task or failure prevalence in deployment.

\begin{wraptable}[23]{r}{0.4\textwidth}
    \centering
    \small
    \caption{\textbf{Composition of PatternEval across three task families}. Subtotals report family-level prompt counts.}    
    \label{tab:task_family_composition}
    \vspace{1mm}

    \renewcommand{\arraystretch}{1.10}
    \setlength{\tabcolsep}{6pt}

    \begin{adjustbox}{max width=\linewidth}
    \begin{tabular}{l r}
        \toprule
        \textbf{Constituent category}
        & \textbf{Number} \\
        \midrule

        \rowcolor{black!10}
        \multicolumn{2}{c}{
            \textbf{VG}\quad
            \textbf{Visual perception and grounding}
        } \\
        OOD perception      & 524 \\
        Content recognition & 350 \\
        Grounding           & 251 \\
        Factuality          & 56  \\
        \multicolumn{1}{l}{-- \textit{Subtotal}}
                            & \textit{1,181} \\
        \addlinespace[2pt]
        \rowcolor{black!10}
        \multicolumn{2}{c}{
            \textbf{OS}\quad
            \textbf{OCR and structured-image understanding}
        } \\
        OCR                 & 299 \\
        Chart understanding & 280 \\
        \multicolumn{1}{l}{-- \textit{Subtotal}}
                            & \textit{579} \\
        \addlinespace[2pt]
        \rowcolor{black!10}
        \multicolumn{2}{c}{
            \textbf{KR}\quad
            \textbf{Multimodal knowledge reasoning}
        } \\
        STEM                & 200 \\
        Knowledge           & 185 \\
        Reasoning           & 270 \\
        \multicolumn{1}{l}{-- \textit{Subtotal}}
                            & \textit{655} \\
        \midrule
        \textbf{Total}      & \textbf{2,415} \\
        \bottomrule
    \end{tabular}
    \end{adjustbox}
\end{wraptable}

\paragraph{PatternEval Composition.}
As summarized in \Cref{tab:task_family_composition}, PatternEval contains 2,415 prompts drawn from nine constituent categories and organized into three task families. \textbf{VG} (visual perception and grounding) is the largest family, comprising 1,181 prompts (48.9\%) across OOD perception, content recognition, grounding, and factuality. These tasks emphasize accurate visual interpretation and faithful grounding of responses in image content, while also covering cases in which familiar recognition patterns may not transfer reliably. \textbf{OS} (OCR and structured-image understanding) contains 579 prompts (24.0\%), spanning both text-centric OCR tasks and chart understanding. This family introduces structure-sensitive inputs for which successful responses require not only extracting local visual elements, but also preserving their spatial, tabular, or quantitative relations. The remaining 655 prompts (27.1\%) constitute \textbf{KR} (multimodal knowledge reasoning), including STEM, knowledge, and general reasoning tasks that require integrating visual evidence with domain knowledge or multi-step inference. Although the three families differ in size, their combination deliberately covers perception-heavy, structure-sensitive, and knowledge-intensive settings. This breadth enables PatternEval to examine whether response-pattern failures are confined to particular task types or persist across heterogeneous multimodal demands under matched reasoning interfaces.

\subsection{Evaluation Protocol}
\begin{figure}[htbp]
    \centering
    \includegraphics[width=0.95\linewidth]{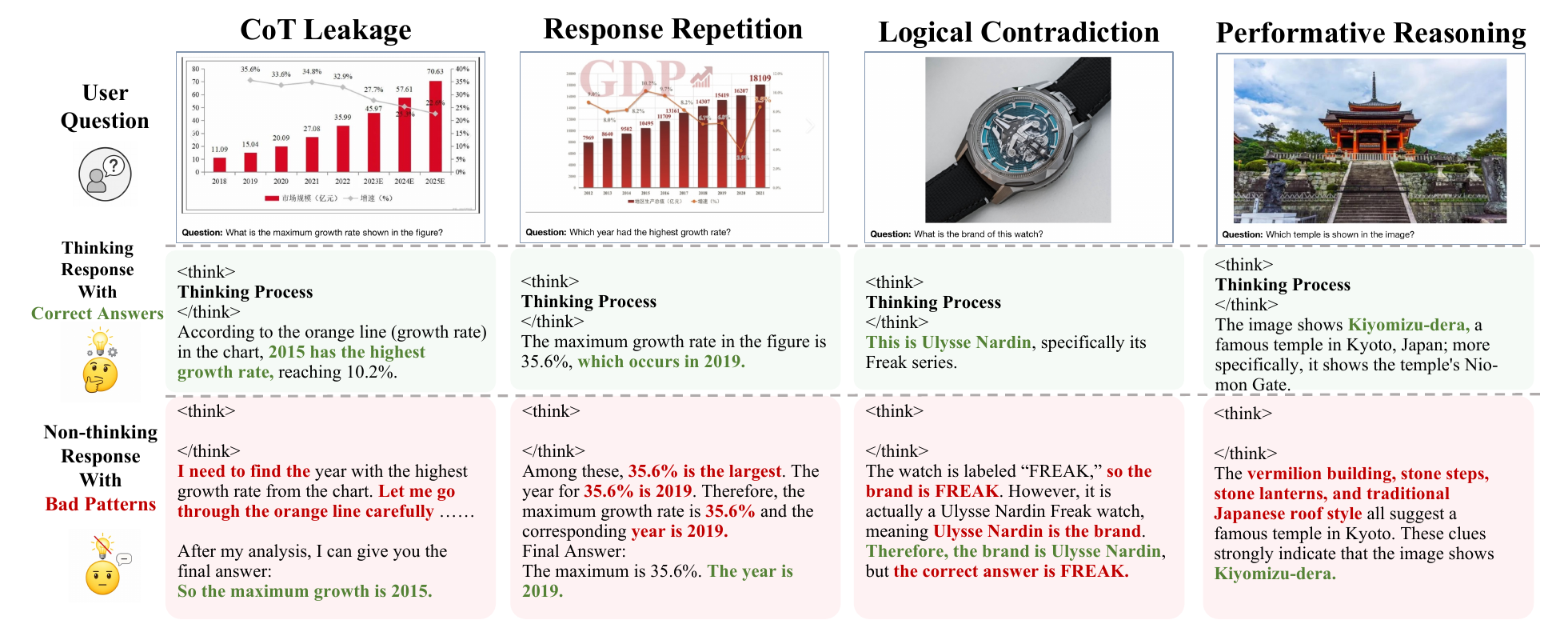}
    \caption{\textbf{Representative response-pattern annotations.} Each column contrasts an acceptable final response with one that triggers the corresponding PatternEval label.}
    \label{fig:bad-pattern-examples}
\end{figure}

\paragraph{Response-pattern taxonomy.}
\label{sec:response-pattern-taxonomy}
PatternEval targets four recurrent failures identified through qualitative inspection of multimodal rollout traces: chain-of-thought leakage, response repetition, logical contradiction, and performative reasoning.
These four failure patterns occur in user-facing responses yet remain difficult for correctness verifiers to detect. We use the following operational definitions:
\begin{itemize}[leftmargin=*]
    \item \textbf{Chain-of-thought leakage (CoT):} internal-process traces, draft analysis, self-correction, or explicit thinking markers appear in the final user-visible answer. A concise, polished explanation is excluded. Because generated rationales need not faithfully reveal a model's internal causal process \citep{turpin2023unfaithful}, this is an observable response-style label rather than evidence that private internal states were exposed.
    \item \textbf{Response repetition (Rep):} the response unnecessarily repeats the same sentence, semantic content, structure, or conclusion in a way that harms efficiency or readability; necessary restatement and limited emphasis are excluded.
    \item \textbf{Logical contradiction (Con):} the final answer retains mutually incompatible claims about the same object, value, or conclusion under the same conditions without resolving the conflict.
    \item \textbf{Performative reasoning (PR):} the response presents an unsupported reasoning wrapper, cites evidence that does not entail its conclusion, or stages analysis without information gain. In multimodal tasks, relevant support should refer to observable objects, text, quantities, positions, relations, chart values, or paths.
\end{itemize}
Taken together, CoT and Rep are the dominant response-pattern failures and primarily capture undesirable user-visible form, whereas Con and PR require semantic and multimodal consistency judgments. The four labels are not mutually exclusive. Because contradictions and unsupported claims often appear within leaked deliberation, we apply a CoT-priority attribution rule: evidence already attributed to CoT is not counted again as Con or PR, while an independent contradiction or unsupported justification outside the leaked reasoning trace may still trigger the corresponding label. Rep may likewise co-occur with CoT when it reflects a distinct repetitive pattern. Consequently, the four pattern rates are coupled and should be interpreted jointly.

\paragraph{Evaluation settings.}
For each image--prompt pair $x=(v,q)$ with reference answer $g$, model $M$ produces an output under inference mode $m\in\{\mathrm{NT},\mathrm{T}\}$:
\begin{equation}
\tilde{y}_m \sim M(\cdot\mid x,m),
\end{equation}
where $\mathrm{NT}$ and $\mathrm{T}$ denote non-thinking and thinking inference, respectively. If the interface exposes separate reasoning and final-response channels, we define $\tilde{y}_m=(r_m,y_m)$, where $r_m$ is the optional reasoning-channel output and $y_m$ is the user-facing response. We evaluate only $y_m$; any separately returned $r_m$ is excluded from both correctness and pattern evaluation. None of the evaluated non-thinking interfaces returns a separate reasoning channel. This notation concerns only observable model outputs and does not assume private internal states.

Each response $y_m$ is independently evaluated by a verifier judge and a pattern judge. The verifier judge assigns
\begin{equation}
c(x,y_m,g)=
\mathcal{J}_{\mathrm{ver}}(x,y_m,g)
\in\{0,1\},
\end{equation}
where $c=1$ denotes a correct answer. We instantiate $\mathcal{J}_{\mathrm{ver}}$ with \textbf{Qwen3-Max} and apply the task-specific evaluation criteria of the corresponding source dataset.

The pattern judge assigns
\begin{equation}
\mathbf{b}(x,y_m)=\mathcal{J}_{\mathrm{pat}}(x,y_m)=
[b_{\mathrm{cot}},b_{\mathrm{rep}},b_{\mathrm{con}},b_{\mathrm{pr}}]
\in\{0,1\}^{4}.
\end{equation}
We instantiate $\mathcal{J}_{\mathrm{pat}}$ with \textbf{Seed-2.0-Pro}. The four labels indicate chain-of-thought leakage, response repetition, logical contradiction, and performative reasoning, respectively, with $b_k=1$ indicating that the corresponding failure is present. The complete pattern-judge prompt is provided in \Cref{app:meta-judge-prompt}.

For a set of $N$ prompts, define the bad pattern trigger indicator as
\begin{equation}
\label{eq:any-failure}
b_{\mathrm{any}}(x,y_m)
=
\mathbf{1}\!\left[\max_k b_k(x,y_m)=1\right].
\end{equation}
We then compute
\begin{equation}
\mathrm{Acc}_m=\frac{1}{N}\sum_{i=1}^{N}c(x_i,y_{i,m},g_i),
\qquad
\mathrm{Trigger}_m=
\frac{1}{N}\sum_{i=1}^{N}
b_{\mathrm{any}}(x_i,y_{i,m}).
\end{equation}
Thus, higher $\mathrm{Acc}$ indicates better task performance, whereas lower $\mathrm{Trigger}$ indicates fewer undesirable user-facing response patterns.

Because both modes are evaluated on the same prompt set, we define their accuracy and response-pattern gaps as
$\Delta_{\mathrm{acc}}=\mathrm{Acc}_{\mathrm{T}}-
\mathrm{Acc}_{\mathrm{NT}}$ and $\Delta_{\mathrm{pat}}=\mathrm{Trigger}_{\mathrm{NT}}-\mathrm{Trigger}_{\mathrm{T}}$. A positive $\Delta_{\mathrm{acc}}$ indicates an accuracy advantage for thinking inference, while positive $\Delta_{\mathrm{pat}}$ indicates a higher aggregate pattern-failure rate under non-thinking inference. The sign of $\Delta_{\mathrm{pat}}$ records the direction of this interface gap, and $|\Delta_{\mathrm{pat}}|$ records its magnitude.
We report both mode-specific Trigger rates together with $\Delta_{\mathrm{pat}}$, since the same gap may arise when both modes have either low or high failure rates.

\subsection{Meta-Judge Analysis}
\label{sec:meta-judge-design}

\paragraph{Pattern-judge calibration-set construction.}
To calibrate the automatic pattern judge, we construct a separately annotated pattern-judge calibration set containing 2,500 fixed responses. These responses are not sampled directly from PatternEval. Instead, we collect rollouts generated by model variants obtained from the supervised fine-tuning (SFT), MixRL, and OPD stages of Hy3-Vision on a separate test set drawn from the same source tasks as PatternEval. This design evaluates every candidate judge on the same responses while covering output distributions produced by different post-training stages and avoiding overlap with the benchmark samples used for model evaluation.

We use Seed-2.0-Pro, Kimi-K2.6, and Qwen3.5-397B as the initial judges. Each judge assigns binary labels for the four failure types. We then divide the rollout responses into consensus samples, for which all three judges produce the same labels, and disputed samples, for which their predictions differ. The two strata are sampled at an approximate $7{:}3$ ratio, respectively, so that the resulting dataset contains both representative cases and difficult cases that distinguish judge capability. During filtering, we additionally control the predicted bad pattern trigger rate at approximately $70\%$, ensuring sufficient positive examples for reliable failure-label-level evaluation. The selected 2,500 responses are subsequently annotated by human annotators under the same four-label rubric, and the human annotations serve as the reference labels.

\paragraph{Meta-judge evaluation and results.}
Response-pattern labels are less mechanically verifiable than final-answer correctness and therefore require an explicitly calibrated judging protocol. We evaluate five candidate judges on the same 2,500 fixed responses against their human reference annotations. As shown in Table~\ref{tab:meta_judge_candidate_scores}, each judge predicts the four binary pattern labels, and we report failure-label-level precision, recall, and F1 together with their macro-averages.
\begin{table}[H]
    \centering
    \scriptsize
    \caption{
        \textbf{Meta-judge calibration (\%).}
        Overall averages over the four failure labels;
        bold and underlined values indicate the best and second-best results, respectively.
    }
    \label{tab:meta_judge_candidate_scores}
    \setlength{\tabcolsep}{2.4pt}
    \renewcommand{\arraystretch}{0.94}
    \begin{adjustbox}{max width=\textwidth,center}
    \begin{tabular}{l!{\color{black!35}\vrule width 0.35pt}l ccc ccc ccc ccc ccc}
        \toprule
        \multirow{2}{*}{\textbf{Judge}}
        & \multirow{2}{*}{\textbf{Input}}
        & \multicolumn{3}{c}{\textbf{CoT.}}
        & \multicolumn{3}{c}{\textbf{Rep.}}
        & \multicolumn{3}{c}{\textbf{Con.}}
        & \multicolumn{3}{c}{\textbf{PR.}}
        & \multicolumn{3}{c}{\textbf{Overall}} \\
        \cmidrule(lr){3-5}
        \cmidrule(lr){6-8}
        \cmidrule(lr){9-11}
        \cmidrule(lr){12-14}
        \cmidrule(lr){15-17}
        & & P & R & F1 & P & R & F1 & P & R & F1 & P & R & F1 & P & R & F1 \\
        \midrule

        \multirow{2}{*}{GPT-5.5}
        & Image+text
        & 98.5
        & \textbf{92.4}
        & \textbf{95.3}
        & 83.5
        & \underline{92.9}
        & \underline{87.9}
        & 46.3
        & \underline{85.5}
        & 60.1
        & 54.4
        & \textbf{79.3}
        & \textbf{64.5}
        & 70.7
        & \textbf{87.5}
        & \underline{77.0} \\

        & Text only
        & 98.4
        & \underline{92.2}
        & \underline{95.2}
        & 83.2
        & \textbf{93.8}
        & \textbf{88.2}
        & 47.5
        & \textbf{88.9}
        & 61.9
        & 55.2
        & \underline{75.0}
        & \underline{63.6}
        & 71.1
        & \textbf{87.5}
        & \textbf{77.2} \\

        \cmidrule(lr){1-2}
        \addlinespace[1.5pt]

        \multirow{2}{*}{Seed-2.0-Pro}
        & Image+text
        & \underline{98.6}
        & 84.8
        & 91.2
        & 83.5
        & 91.5
        & 87.3
        & 53.7
        & 74.8
        & \textbf{62.5}
        & 50.4
        & 74.1
        & 60.0
        & 71.6
        & \underline{81.3}
        & 75.3 \\

        & Text only
        & \underline{98.6}
        & 84.7
        & 91.1
        & 83.4
        & 91.6
        & 87.3
        & \underline{55.6}
        & 67.5
        & 61.0
        & 44.8
        & 68.1
        & 54.0
        & 70.6
        & 78.0
        & 73.4 \\

        \cmidrule(lr){1-2}
        \addlinespace[1.5pt]

        Hy3
        & Text only
        & \textbf{98.7}
        & 83.0
        & 90.2
        & 84.2
        & 80.7
        & 82.4
        & 48.8
        & 69.8
        & 57.4
        & 49.5
        & 57.1
        & 53.0
        & 70.3
        & 72.7
        & 70.8 \\

        \cmidrule(lr){1-2}
        \addlinespace[1.5pt]

        \multirow{2}{*}{Kimi-K2.6}
        & Image+text
        & 97.7
        & 89.1
        & 93.2
        & \textbf{87.1}
        & 76.8
        & 81.6
        & \textbf{55.8}
        & 70.1
        & 62.1
        & \underline{58.4}
        & 40.7
        & 48.0
        & \textbf{74.8}
        & 69.2
        & 71.2 \\

        & Text only
        & 97.7
        & 87.7
        & 92.4
        & \underline{86.5}
        & 79.2
        & 82.7
        & 49.4
        & 65.0
        & 56.1
        & \textbf{64.1}
        & 36.0
        & 46.1
        & \underline{74.4}
        & 67.0
        & 69.3 \\

        \cmidrule(lr){1-2}
        \addlinespace[1.5pt]

        \multirow{2}{*}{Qwen3.5-397B}
        & Image+text
        & 97.6
        & 87.3
        & 92.2
        & 86.4
        & 84.6
        & 85.5
        & 44.7
        & 75.2
        & 56.1
        & 36.2
        & 61.6
        & 45.6
        & 66.2
        & 77.2
        & 69.9 \\

        & Text only
        & 97.5
        & 88.5
        & 92.8
        & 83.2
        & 88.6
        & 85.8
        & 52.0
        & 77.8
        & \underline{62.3}
        & 55.6
        & 33.5
        & 41.8
        & 72.1
        & 72.1
        & 70.7 \\

        \bottomrule
    \end{tabular}
    \end{adjustbox}
\end{table}

As shown in \Cref{tab:meta_judge_candidate_scores}, CoT leakage and repetition are judged more reliably because they often exhibit explicit surface-level cues, whereas contradiction and performative reasoning require finer-grained semantic assessment, including cross-sentence consistency, visual grounding, and distinguishing evidence-supported analysis from merely reasoning-like language. Accordingly, the latter two labels yield lower and more variable precision and recall across judges. Since the operational judge is used for both benchmark evaluation and training-data filtering, we consider recall alongside precision and F1 to reduce undetected response-pattern failures. GPT-5.5 achieves the highest overall F1, while Seed-2.0-Pro with image access attains the best contradiction F1 and directly grounds its decisions in the visual input. Balancing failure-label-level performance, multimodal grounding, recall, and inference cost, we select \textbf{Seed-2.0-Pro with image access} as the operational pattern judge for PatternEval.

\section{Experiments}
\label{sec:experiments}

\subsection{Experimental Setup}
\label{sec:experimental-setup}

\paragraph{Models.}
We evaluate 25 model configurations spanning the Qwen series, Kimi, Seed, Claude, Mimo, and GPT families. Each configuration is evaluated under matched thinking and non-thinking modes. We follow the official default inference settings and vary only the available reasoning control, while keeping the remaining exposed decoding parameters fixed. For externally hosted models, we use the reasoning modes or effort controls provided by their APIs. Full nine-constituent-category results are reported in \Cref{app:more-results}.

\paragraph{Protocol.}
For each configuration, we generate thinking and non-thinking responses on the same 2,415 PatternEval prompts, enabling paired comparisons across inference modes. \textbf{Qwen3-Max} evaluates task correctness according to the protocol and reference answers of each source task, whereas \textbf{Seed-2.0-Pro}, with access to the input image, evaluates the four response patterns. We report PatternEval Acc, Trigger, the per-label failure rates for CoT, Rep, Con, and PR, the VG, OS, and KR task-family Trigger rates, and the matched-mode differences $\Delta_{\mathrm{acc}}$ and $\Delta_{\mathrm{pat}}$.

\subsection{Main Results}
\label{sec:thinking-nonthinking-patterns}

\Cref{tab:detailed_pattern_results} reports the PatternEval results across models under thinking and non-thinking inference. Based on these results, we draw the following key takeaways.


\paragraph{Takeaway 1: Response-pattern gaps persist across model families on PatternEval.}
Every evaluated pair has a positive response-pattern gap across open- and closed-source models, dense and mixture-of-experts architectures, and parameter scales. Although several frontier models substantially reduce the gap between inference modes, $\Delta_{\mathrm{pat}}$ exceeds 20 percentage points for 17 of the 25 pairs. This cross-model consistency shows that stronger task capability alone does not ensure response-pattern alignment under difficult conditions.

\paragraph{Takeaway 2: Non-thinking inference often exposes deliberation-style content.}
CoT leakage is the most prominent failure mode and constitutes the dominant source of bad-pattern triggers. Under our operational definition, non-thinking responses may contain process narration, redundant intermediate analysis, self-correction, or repeated reasoning fragments in the user-visible answer. The label characterizes observable response form independently of any claim about a model's private internal computation.

\paragraph{Takeaway 3: Pattern failures are unevenly distributed.}
Taking the arithmetic mean over the 50 model--mode rows, CoT leakage and response repetition have marginal trigger rates of 12.75\% and 8.49\%, respectively, compared with 4.72\% for performative reasoning and 2.74\% for logical contradiction. This imbalance is more pronounced under non-thinking inference. Performative reasoning and logical contradiction are detected less often and may also co-occur with more salient failures such as CoT leakage; because the taxonomy uses a CoT-priority attribution rule, the marginal category rates should not be interpreted as independent prevalence estimates.

\begin{table}[H]
    \centering
    \scriptsize
    \caption{
    \textbf{Performance and response-pattern failures under thinking and non-thinking inference.}
    All values are percentages; signed gaps are percentage points.
    Light-blue and light-orange cells denote the best and worst results,
    respectively, computed separately within each inference mode.
    }
    \vspace{1mm}
    \label{tab:detailed_pattern_results}

    \begin{adjustbox}{width=0.95\textwidth}
    \begin{tabular}{l c cc cc cccc ccc}
        \toprule
        \multirow{2}{*}{\textbf{Model}}
        & \multirow{2}{*}{\textbf{Mode}}
        & \multicolumn{2}{c}{\textbf{Mode Gap}}
        & \multicolumn{2}{c}{\textbf{Overall Results}}
        & \multicolumn{4}{c}{\textbf{Pattern Failures}}
        & \multicolumn{3}{c}{\textbf{Task-family Trigger}} \\
        \cmidrule(lr){3-4}
        \cmidrule(lr){5-6}
        \cmidrule(lr){7-10}
        \cmidrule(lr){11-13}
        & & $\Delta_{\mathrm{acc}}$ & $\Delta_{\mathrm{pat}}$
          & Acc$\uparrow$ & Trigger$\downarrow$
          & CoT$\downarrow$ & Rep$\downarrow$ & Con$\downarrow$ & PR$\downarrow$
          & VG$\downarrow$ & OS$\downarrow$ & KR$\downarrow$ \\
        \midrule

        \multirow{2}{*}{Qwen3.5-0.8B}
        & \NTicon & \multirow{2}{*}{-2.09} & \multirow{2}{*}{6.95}
        & \worstcell{25.22} & \worstcell{75.94}
        & 32.22 & \worstcell{38.92} & \worstcell{26.87}
        & \worstcell{36.81}
        & \worstcell{84.32} & \worstcell{50.34}
        & \worstcell{83.51} \\
        & \Ticon & & &
        \worstcell{23.13} & \worstcell{68.99}
        & \worstcell{13.50} & \worstcell{29.94}
        & \worstcell{34.49} & \worstcell{47.83}
        & \worstcell{75.08} & \worstcell{53.62}
        & \worstcell{71.60} \\

        \addlinespace[0.5pt]
        \multirow{2}{*}{Qwen3.5-4B}
        & \NTicon & \multirow{2}{*}{6.37} & \multirow{2}{*}{24.23}
        & 45.96 & 46.42 & 28.24 & 25.55 & 7.62 & 10.60
        & 50.68 & 19.31 & 62.75 \\
        & \Ticon & & &
        52.33 & 22.19 & 11.18 & 10.60 & 3.11 & 6.50
        & 25.51 & 8.45 & 28.40 \\

        \addlinespace[0.5pt]
        \multirow{2}{*}{Qwen3.5-9B}
        & \NTicon & \multirow{2}{*}{9.05} & \multirow{2}{*}{21.36}
        & 50.15 & 36.19 & 25.22 & 19.05 & 5.13 & 5.01
        & 40.17 & 13.45 & 49.16 \\
        & \Ticon & & &
        59.20 & 14.83 & 8.41 & 6.30 & 1.45 & 3.44
        & 16.95 & 4.31 & 20.34 \\

        \addlinespace[0.5pt]
        \multirow{2}{*}{Qwen3.5-27B}
        & \NTicon & \multirow{2}{*}{4.98} & \multirow{2}{*}{24.01}
        & 59.07 & 29.32 & 23.44 & 15.65 & 0.91 & 2.48
        & 30.93 & 7.41 & 45.80 \\
        & \Ticon & & &
        64.05 & 5.31 & 2.94 & 1.78 & 0.17 & 1.16
        & 5.51 & 1.21 & 8.58 \\

        \addlinespace[0.5pt]
        \multirow{2}{*}{Qwen3.5-35B-A3B}
        & \NTicon & \multirow{2}{*}{8.05} & \multirow{2}{*}{26.63}
        & 57.02 & 33.00 & 25.13 & 17.97 & 2.48 & 3.02
        & 36.02 & 12.59 & 45.65 \\
        & \Ticon & & &
        65.07 & 6.37 & 3.87 & 2.50 & 0.62 & 1.08
        & 7.94 & 1.55 & 7.86 \\

        \addlinespace[0.5pt]
        \multirow{2}{*}{Qwen3.5-122B-A10B}
        & \NTicon & \multirow{2}{*}{7.10} & \multirow{2}{*}{22.59}
        & 59.20 & 28.45 & 23.23 & 15.36 & 1.08 & 1.74
        & 30.85 & 8.62 & 41.68 \\
        & \Ticon & & &
        66.30 & 5.86 & 4.18 & 2.09 & 0.17 & 1.05
        & 5.16 & 0.86 & 11.57 \\

        \addlinespace[0.5pt]
        \multirow{2}{*}{Qwen3.5-397B-A17B}
        & \NTicon & \multirow{2}{*}{6.97} & \multirow{2}{*}{25.09}
        & 61.96 & 29.32 & 25.71 & 15.65 & 0.66 & 1.12
        & 30.51 & 9.48 & 44.73 \\
        & \Ticon & & &
        68.93 & 4.23 & 2.98 & 1.12 & 0.17 & 0.83
        & 4.75 & 0.69 & 6.42 \\

        \midrule

        \multirow{2}{*}{Qwen3.6-27B}
        & \NTicon & \multirow{2}{*}{6.39} & \multirow{2}{*}{26.88}
        & 57.33 & 32.81 & 26.14 & 18.31 & 1.66 & 2.57
        & 35.54 & 10.69 & 47.48 \\
        & \Ticon & & &
        63.72 & 5.93 & 3.69 & 2.28 & 0.33 & 1.08
        & 6.80 & 1.55 & 8.24 \\

        \addlinespace[0.5pt]
        \multirow{2}{*}{Qwen3.6-35B-A3B}
        & \NTicon & \multirow{2}{*}{7.68} & \multirow{2}{*}{26.90}
        & 56.22 & 32.80 & 26.25 & 17.27 & 2.36 & 2.15
        & 37.12 & 10.34 & 44.89 \\
        & \Ticon & & &
        63.90 & 5.90 & 3.58 & 1.91 & 0.42 & 1.41
        & 6.12 & 1.21 & 9.71 \\

        \addlinespace[0.5pt]
        \multirow{2}{*}{Qwen3.7-Max}
        & \NTicon & \multirow{2}{*}{5.23} & \multirow{2}{*}{21.21}
        & 67.07 & 26.77 & 22.28 & 16.79 & 0.46 & 0.58
        & 29.69 & 13.26 & 33.33 \\
        & \Ticon & & &
        72.30 & 5.56 & 3.98 & 2.70 & 0.17 & 0.75
        & 6.54 & 2.25 & 6.73 \\

        \addlinespace[0.5pt]
        \multirow{2}{*}{Qwen3.7-Plus}
        & \NTicon & \multirow{2}{*}{4.69} & \multirow{2}{*}{21.23}
        & 66.28 & 27.82 & 23.13 & 17.50 & 0.58 & 0.66
        & 29.24 & 13.99 & 37.52 \\
        & \Ticon & & &
        70.97 & 6.59 & 4.85 & 3.90 & 0.21 & 0.70
        & 7.63 & 1.73 & 9.04 \\

        \midrule

        \multirow{2}{*}{Qwen3-VL-30B-A3B}
        & \NTicon & \multirow{2}{*}{0.66} & \multirow{2}{*}{30.88}
        & 48.21 & 41.54 & 32.03 & 20.43 & 3.58 & 4.22
        & 42.11 & 24.66 & 55.62 \\
        & \Ticon & & &
        48.87 & 10.66 & 0.54 & 2.90 & 2.37 & 6.60
        & 15.15 & 2.93 & 9.47 \\

        \addlinespace[0.5pt]
        \multirow{2}{*}{Qwen3-VL-32B}
        & \NTicon & \multirow{2}{*}{2.54} & \multirow{2}{*}{30.21}
        & 53.21 & 35.03 & 25.83 & 21.31 & 1.49 & 2.07
        & 41.09 & 12.76 & 43.88 \\
        & \Ticon & & &
        55.75 & 4.82 & 0.35 & 1.53 & 1.10 & 2.63
        & 6.27 & 1.38 & 5.53 \\

        \midrule

        \multirow{2}{*}{Kimi-K2.5}
        & \NTicon & \multirow{2}{*}{5.73} & \multirow{2}{*}{38.29}
        & 61.33 & 44.43 & 42.07 & 11.59
        & \bestcell{0.37} & 0.91
        & 44.24 & 24.14 & 62.75 \\
        & \Ticon & & &
        67.06 & 6.14 & 3.55 & 1.00 & 1.09 & 1.34
        & 6.69 & 2.76 & 8.19 \\

        \addlinespace[0.5pt]
        \multirow{2}{*}{Kimi-K2.6}
        & \NTicon & \multirow{2}{*}{6.72} & \multirow{2}{*}{48.64}
        & 66.47 & 49.63 & \worstcell{47.39} & 16.74
        & 0.54 & \bestcell{0.50}
        & 47.92 & 28.45 & 71.45 \\
        & \Ticon & & &
        73.19 & 0.99 & 0.13 & \bestcell{0.13}
        & 0.27 & 0.49
        & 1.41 & \bestcell{0.34} & 0.85 \\

        \midrule

        \multirow{2}{*}{Seed-1.6-Vision}
        & \NTicon & \multirow{2}{*}{2.24} & \multirow{2}{*}{20.35}
        & 59.49 & 29.63 & 24.28 & 20.77 & 0.71 & 3.34
        & 41.38 & 10.52 & 25.31 \\
        & \Ticon & & &
        61.73 & 9.28 & 2.84 & 3.55 & 1.63 & 4.05
        & 14.10 & 3.10 & 5.97 \\

        \addlinespace[0.5pt]
        \multirow{2}{*}{Seed-2.0-Pro}
        & \NTicon & \multirow{2}{*}{6.13} & \multirow{2}{*}{3.81}
        & 66.85 & 6.94 & 4.51 & 0.75 & 1.34 & 1.04
        & 7.31 & 3.62 & 9.28 \\
        & \Ticon & & &
        72.98 & 3.13 & 2.09 & 0.25 & 0.21 & 0.67
        & 3.91 & 0.86 & 3.77 \\

        \addlinespace[0.5pt]
        \multirow{2}{*}{Seed-2.1-Pro}
        & \NTicon & \multirow{2}{*}{13.10} & \multirow{2}{*}{9.06}
        & \bestcell{70.44} & 11.37
        & 8.90 & 3.05 & 1.25 & 0.54
        & 12.83 & 3.45 & 15.88 \\
        & \Ticon & & &
        \bestcell{83.54} & 2.31
        & 1.72 & 0.34 & 0.05 & \bestcell{0.39}
        & 2.14 & 2.24 & 2.67 \\

        \midrule

        \multirow{2}{*}{Claude-Opus-4.6}
        & \NTicon & \multirow{2}{*}{8.18} & \multirow{2}{*}{29.59}
        & 59.45 & 33.18 & 30.35 & 8.79 & 1.23 & 1.27
        & 39.48 & 5.37 & 47.12 \\
        & \Ticon & & &
        67.63 & 3.59 & 1.45 & 0.23 & 0.56 & 1.49
        & 4.83 & 1.22 & 3.84 \\

        \addlinespace[0.5pt]
        \multirow{2}{*}{Claude-Opus-4.8}
        & \NTicon & \multirow{2}{*}{6.58} & \multirow{2}{*}{18.51}
        & 61.12 & 22.06 & 18.93 & 5.11 & 1.18 & 0.89
        & 29.01 & 4.68 & 25.16 \\
        & \Ticon & & &
        67.70 & 3.55 & 1.93 & 0.48 & 0.44 & 1.01
        & 4.24 & 1.56 & 4.18 \\

        \midrule

        \multirow{2}{*}{Mimo-v2-Omni}
        & \NTicon & \multirow{2}{*}{0.31} & \multirow{2}{*}{12.83}
        & 57.21 & 22.85 & 16.37 & 3.74 & 3.66 & 4.03
        & 27.23 & 14.19 & 22.63 \\
        & \Ticon & & &
        57.52 & 10.02 & 3.50 & 1.30 & 1.94 & 5.10
        & 13.99 & 2.71 & 9.48 \\

        \addlinespace[0.5pt]
        \multirow{2}{*}{Mimo-v2.5}
        & \NTicon & \multirow{2}{*}{0.42} & \multirow{2}{*}{13.88}
        & 55.03 & 24.92 & 16.47 & 3.87 & 5.03 & 4.83
        & 34.22 & 8.65 & 22.63 \\
        & \Ticon & & &
        55.45 & 11.04 & 3.41 & 2.34 & 2.13 & 6.52
        & 15.70 & 2.29 & 10.47 \\

        \midrule

        \multirow{2}{*}{GPT-5 Nano}
        & \NTicon & \multirow{2}{*}{13.64} & \multirow{2}{*}{23.14}
        & 30.72 & 35.67 & 2.12 & 7.38 & 9.83 & 28.20
        & 40.92 & 22.97 & 37.46 \\
        & \Ticon & & &
        44.36 & 12.53 & 0.17 & 1.16 & 1.08 & 11.20
        & 19.78 & 2.42 & 8.41 \\

        \addlinespace[0.5pt]
        \multirow{2}{*}{GPT-5.4}
        & \NTicon & \multirow{2}{*}{8.88} & \multirow{2}{*}{2.86}
        & 60.66 & 5.34 & 1.20 & 0.91 & 0.79 & 2.82
        & \bestcell{5.68} & \bestcell{1.21} & 8.40 \\
        & \Ticon & & &
        69.54 & 2.48
        & \bestcell{0.00} & 0.62 & \bestcell{0.04} & 1.82
        & 3.56 & 0.86 & 1.98 \\

        \addlinespace[0.5pt]
        \multirow{2}{*}{GPT-5.5}
        & \NTicon & \multirow{2}{*}{15.70} & \multirow{2}{*}{4.26}
        & 59.20 & \bestcell{5.04}
        & \bestcell{0.29} & \bestcell{0.42} & 0.75 & 4.00
        & 7.32 & \bestcell{1.21} & \bestcell{4.31} \\
        & \Ticon & & &
        74.90 & \bestcell{0.78}
        & \bestcell{0.00} & 0.18 & 0.14 & 0.46
        & \bestcell{1.17} & 0.52 & \bestcell{0.34} \\

        \bottomrule
    \end{tabular}
    \end{adjustbox}
\end{table}

\subsection{Further Analysis}
\label{sec:further-analysis}

We further investigate two factors that may shape the observed response-pattern gap: model capability and response length. Specifically, we examine whether stronger models narrow the discrepancy between thinking and non-thinking inference, and how response length interacts with inference mode and answer correctness in determining pattern-failure incidence.

\begin{figure}[htbp]
    \centering
    \includegraphics[width=0.95\textwidth]{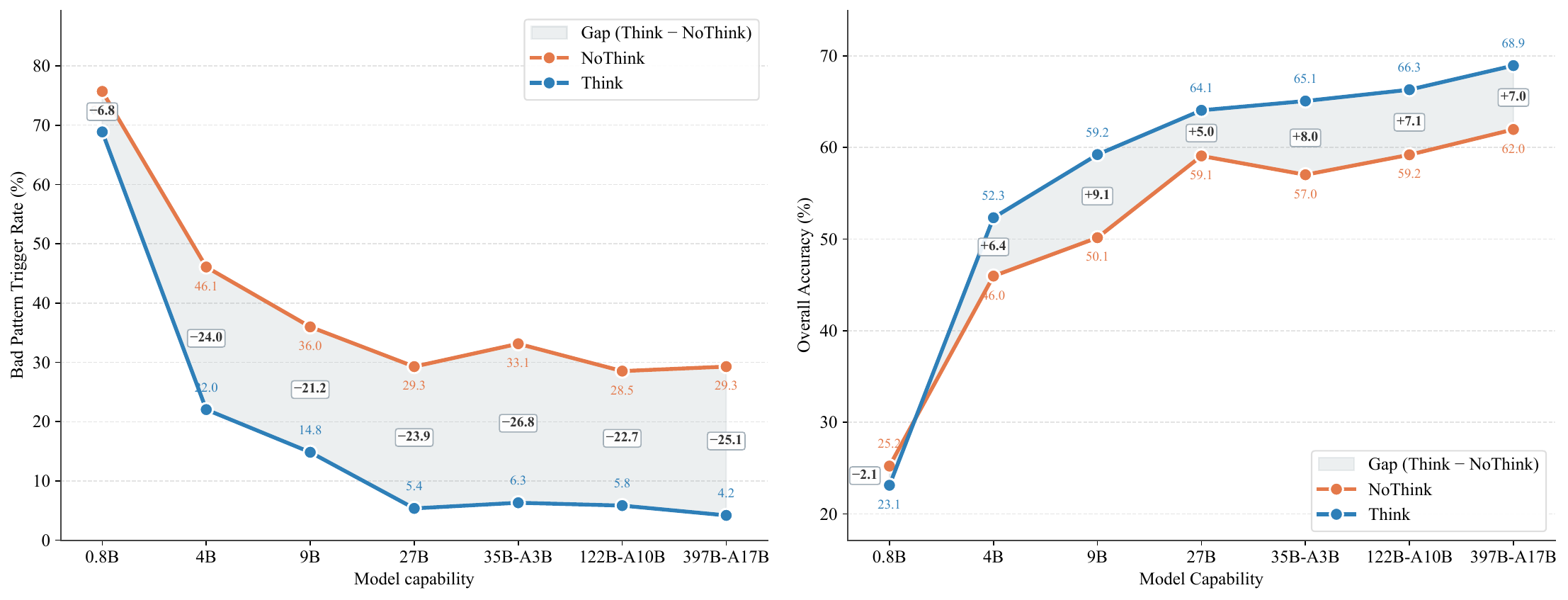}
    \caption{
    \textbf{Model capability and response-pattern failures across the Qwen3.5 series.}
    Shaded annotations report the matched thinking--non-thinking gaps in percentage points.
    }
    \label{fig:qwen-series-capability}
\end{figure}

\paragraph{Model capability.}
The Qwen3.5 series shows that higher PatternEval Acc within a model family is not accompanied by a smaller cross-mode Trigger gap. From 4B to 397B-A17B, PatternEval Acc increases from 45.96\% to 61.96\% under non-thinking inference and from 52.33\% to 68.93\% under thinking inference. Over the same range, thinking-mode Trigger falls from 22.19\% to 4.23\%. Non-thinking Trigger initially decreases from 46.42\% at 4B to 29.32\% at 27B, but then remains between 28.45\% and 33.00\% among the larger models. Consequently, the cross-mode Trigger gap stays between 21.36 and 26.63 percentage points from 4B onward. Within this series, scale, PatternEval Acc, and the cross-mode Trigger gap follow distinct descriptive trends.

\begin{wrapfigure}[20]{r}{0.4\textwidth}
    \centering
    \vspace{-4mm}
    
    \includegraphics[
        width=\linewidth
    ]{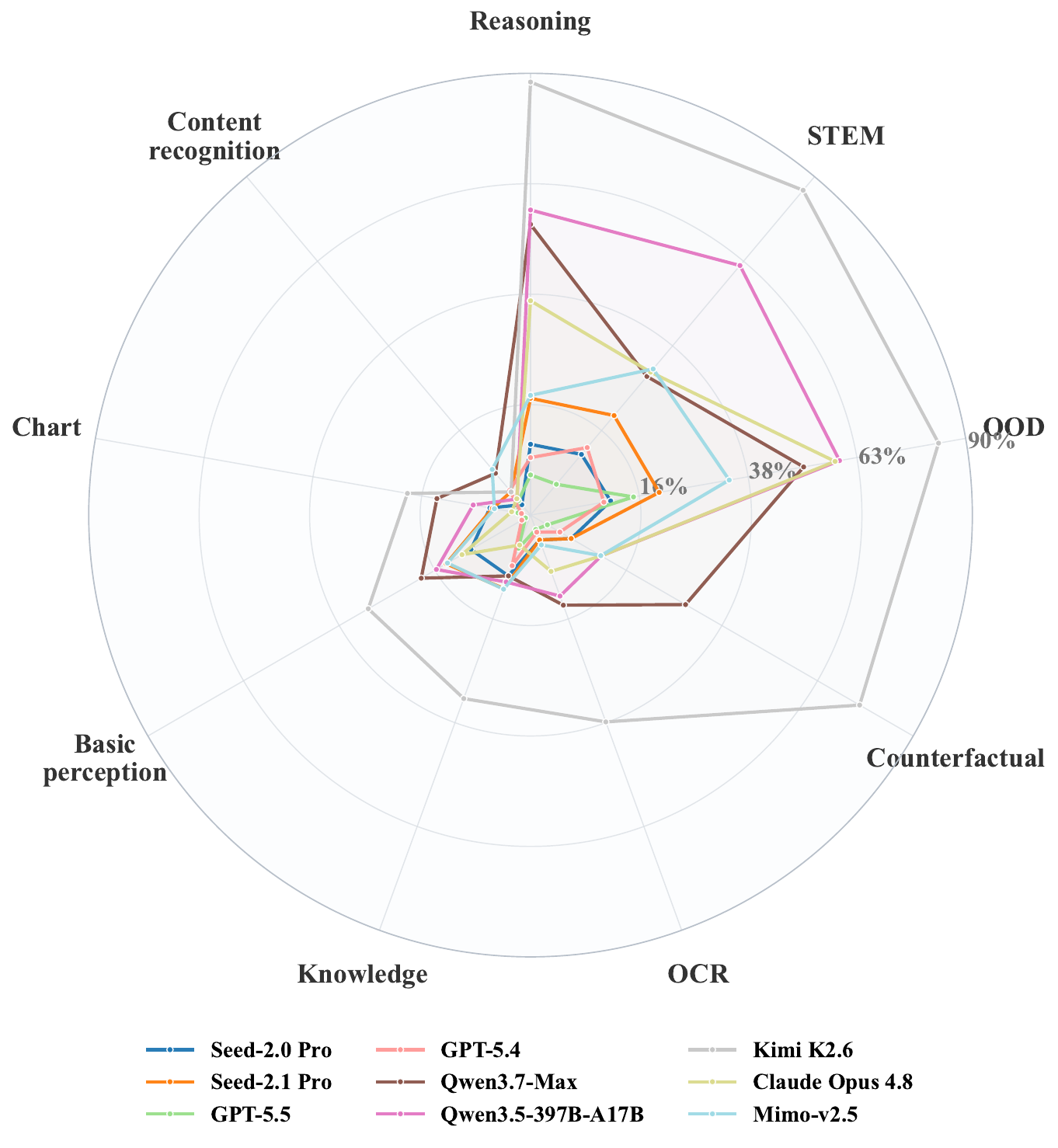}
    
    \caption{
        \textbf{Constituent-category susceptibility under non-thinking.}
        Each polygon reports one model's Trigger across nine constituent task categories.
    }
    \label{fig:nothink-pattern-radar}
    
    \vspace{-2mm}
\end{wrapfigure}

\paragraph{Failure-prone constituent categories.}
The radar in \Cref{fig:nothink-pattern-radar} provides a category-level view of non-thinking Trigger across the nine constituent task categories. Reasoning, STEM, and OOD perception form some of the largest peaks for multiple models, whereas content recognition and chart understanding generally exhibit lower trigger rates. This tendency is not uniform, however, and the relative ordering of categories varies substantially across models. In particular, models with similar aggregate Trigger can arrive at that average through different profiles: failures may be concentrated in a small number of highly susceptible categories or distributed more broadly across the benchmark. Conversely, models with different overall Trigger can still show comparable vulnerability on individual categories. These differences suggest that a single aggregate score may obscure localized response-pattern weaknesses and motivate reporting category-level results alongside the overall metric.

\begin{figure}[htbp]
    \centering
    \includegraphics[width=0.95\textwidth]{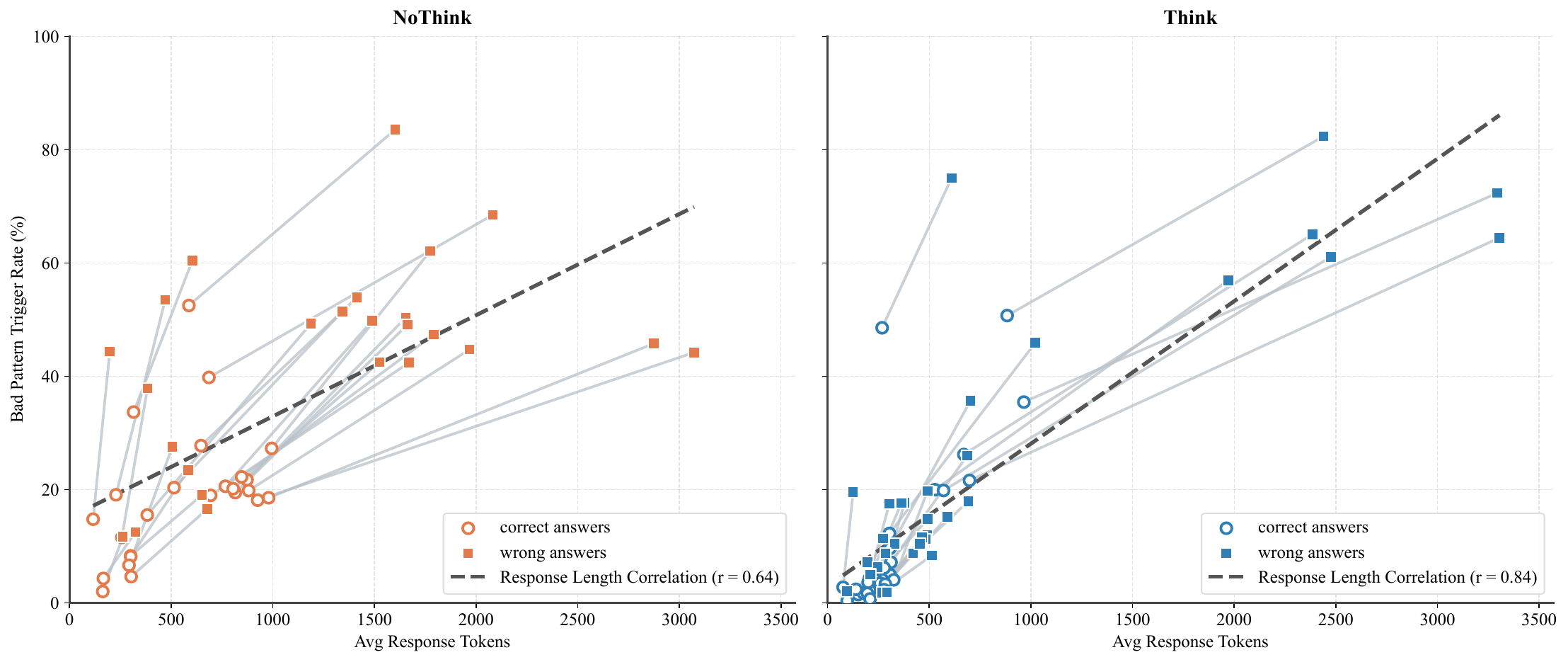}
    \caption{
    \textbf{Model-level response length versus Trigger.}
    Connected markers pair correct and incorrect aggregates; dashed lines show mode-specific correlations.
    }
    \label{fig:response-token-badpattern}
\end{figure}

\paragraph{Response length, correctness, and Trigger.}
\Cref{fig:response-token-badpattern} characterizes the relationship between response length and Trigger using model-by-correctness aggregates. Across the plotted points, average response length is positively associated with Trigger under both inference settings, with Pearson correlations of $r=0.64$ in non-thinking mode and $r=0.84$ in thinking mode. Thus, aggregates containing longer responses also tend to exhibit more user-visible response-pattern failures, with this relationship appearing stronger under thinking inference. This difference is descriptive rather than causal: response length may reflect other factors, such as task difficulty, uncertainty, repetition, or extended but unsuccessful reasoning, rather than directly producing the observed failures. The connected correct--incorrect points provide a complementary within-model comparison. For the same model and inference mode, the incorrect-response aggregate is generally both longer and more failure-prone than the corresponding correct-response aggregate, suggesting that the overall trend is not driven solely by differences in typical response length across models.

\begin{wrapfigure}[20]{r}{0.4\textwidth}
    \centering
    \vspace{-4mm}
    \includegraphics[
        width=\linewidth
    ]{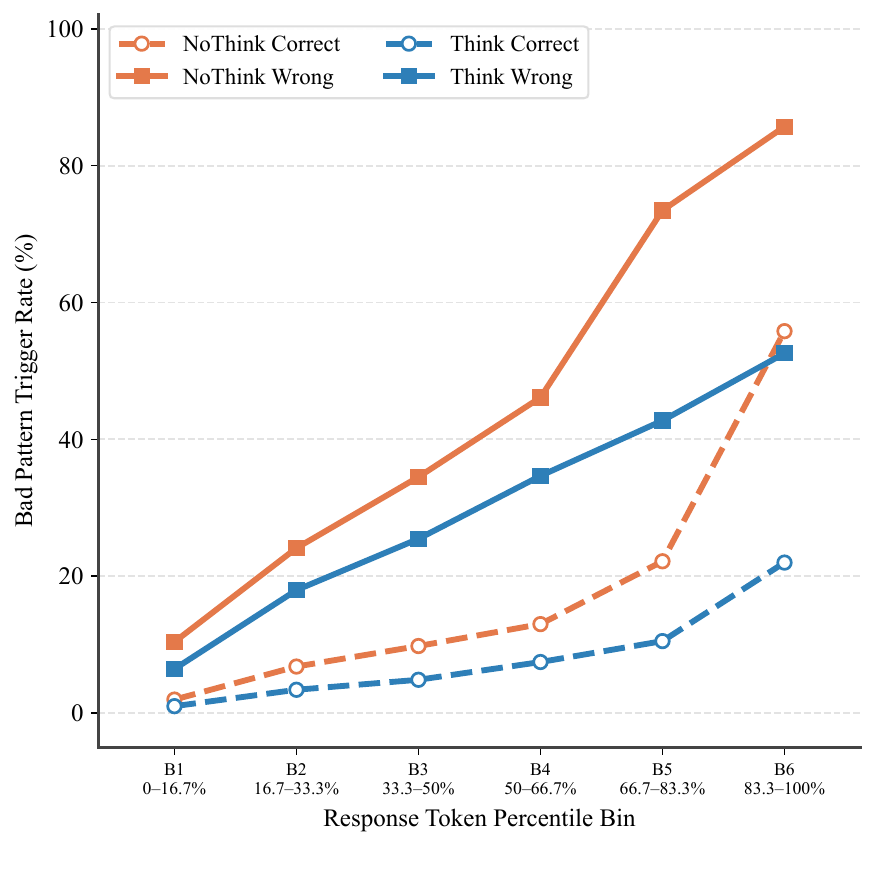}
    \caption{
        \textbf{Trigger across response-length sextiles.}
        Line style denotes correctness and color denotes inference mode.
    }
    \label{fig:response-token-sextiles}
    
    \vspace{-2mm}
\end{wrapfigure}
\Cref{fig:response-token-sextiles} complements this aggregate view by unfolding the response-level length distribution. After responses are divided into six length-based bins, Trigger increases monotonically or near-monotonically across all four mode--correctness groups, indicating that the model-level association is not driven solely by a few unusually verbose models. Within comparable length ranges, incorrect responses remain more failure-prone than correct responses, while non-thinking responses consistently exhibit higher Trigger than thinking responses. The separation is largest in the longest sextile, where Trigger reaches approximately $86\%$ and $56\%$ for incorrect and correct non-thinking responses, compared with $52\%$ and $22\%$ under thinking inference. The high rate among long but correct non-thinking responses is particularly notable, showing that answer correctness alone does not guarantee a well-aligned delivered response. Together, the two views identify long, incorrect non-thinking responses as the highest-observed Trigger regime, while indicating that response length, correctness, and inference mode retain distinct associations with pattern failures.


\FloatBarrier

\section{Pattern-Aware Post-Training}
\label{sec:patternrm}

On PatternEval, all 25 model configurations have higher non-thinking Trigger point estimates. To target these failures during reinforcement learning, we train PatternRM, a reward model that approximates the pattern judge, and optimize the non-thinking policy using complementary signals from the task verifier and PatternRM. This objective preserves the primary correctness reward while directly penalizing the four operational PatternEval labels.

\subsection{PatternRM Training}

\paragraph{Training-data construction.}
We construct the PatternRM supervision corpus from a heterogeneous pool of 90K responses, comprising 60K responses drawn from four existing rollout collections and 30K re-rollouts generated by three additional policies. Each response is independently annotated by Kimi-K2.6, Seed-2.0-Pro, and Qwen3.5-397B with a four-dimensional binary label indicating the presence of CoT leakage, response repetition, logical contradiction, and performative reasoning. To improve annotation reliability, we retain only instances for which all three judges agree on the complete four-label vector, yielding 52,344 unique examples. We then oversample 5,234 selected instances to improve the balance of the training mixture, resulting in 57,578 SFT instances. Among them, 34,547 use thinking-format judgment targets and 23,031 use direct non-thinking targets. Additional details on data sources, sampling, filtering, and target formatting are provided in \Cref{app:patternrl-supervision-corpus}.

\paragraph{Model training.}
We initialize PatternRM from Qwen3.5-27B and train it through supervised fine-tuning to predict the four response-pattern labels. Since PatternRM is repeatedly invoked to provide reward signals during reinforcement learning, inference efficiency is important. We therefore adopt a text-only configuration that evaluates the generated response without taking the associated image as input.

\paragraph{Performance evaluation.}
We evaluate PatternRM on the pattern-judge calibration set. As shown in \Cref{tab:patternrm}, direct-prediction PatternRM attains the highest macro-F1 among the PatternRM variants (71.3\% versus 70.2\%) and avoids the additional decoding required by its thinking counterpart. This comparison reflects predictive performance and decoding cost rather than measured end-to-end latency. Since PatternRM repeatedly scores rollouts during reinforcement learning, we adopt the direct-prediction variant as the reward model for PatternRL.

\begin{table}[H]
    \centering
    \scriptsize
    \caption{
        \textbf{PatternRM judge performance comparison (\%).}
        Bold and underline denote the best and second-best results among non-seed judges, respectively.
    }
    \label{tab:patternrm}
    \setlength{\tabcolsep}{2.4pt}
    \renewcommand{\arraystretch}{0.94}
    \begin{adjustbox}{max width=\textwidth,center}
    \begin{tabular}{l!{\color{black!35}\vrule width 0.35pt}l ccc ccc ccc ccc ccc}
        \toprule
        \multirow{2}{*}{\textbf{Judge}}
        & \multirow{2}{*}{\textbf{Input}}
        & \multicolumn{3}{c}{\textbf{CoT.}}
        & \multicolumn{3}{c}{\textbf{Rep.}}
        & \multicolumn{3}{c}{\textbf{Con.}}
        & \multicolumn{3}{c}{\textbf{PR.}}
        & \multicolumn{3}{c}{\textbf{Overall}} \\
        \cmidrule(lr){3-5}
        \cmidrule(lr){6-8}
        \cmidrule(lr){9-11}
        \cmidrule(lr){12-14}
        \cmidrule(lr){15-17}
        & & P & R & F1 & P & R & F1 & P & R & F1 & P & R & F1 & P & R & F1 \\
        \midrule

        \multirow{2}{*}{Seed-2.0-Pro}
        & Image+text
        & 98.6 & 84.8 & 91.2
        & 83.5 & 91.5 & 87.3
        & 53.7 & 74.8 & 62.5
        & 50.4 & 74.1 & 60.0
        & 71.6 & 81.3 & 75.3 \\

        & Text only
        & 98.6 & 84.7 & 91.1
        & 83.4 & 91.6 & 87.3
        & 55.6 & 67.5 & 61.0
        & 44.8 & 68.1 & 54.0
        & 70.6 & 78.0 & 73.4 \\

        \midrule

        \multirow{2}{*}{Qwen3.5-27B, thinking}
        & Image+text
        & 97.0 & 88.3 & 92.4
        & 81.3 & 88.1 & \textbf{84.6}
        & 45.7 & \textbf{75.5} & 56.9
        & 48.0 & 48.0 & \underline{48.0}
        & 68.0 & \underline{75.0} & \underline{70.5} \\

        & Text only
        & 97.4 & \underline{88.6} & \underline{92.8}
        & 79.2 & \underline{88.2} & 83.5
        & 53.8 & \underline{66.7} & \textbf{59.5}
        & \textbf{63.2} & 29.3 & 40.0
        & 73.4 & 68.2 & 69.0 \\

        \midrule

        \multirow{2}{*}{Qwen3.5-27B, direct}
        & Image+text
        & \underline{97.8} & 86.4 & 91.7
        & 75.8 & 79.6 & 77.7
        & 33.9 & 48.7 & 40.0
        & 29.4 & \underline{72.6} & 41.8
        & 59.2 & 71.8 & 62.8 \\

        & Text only
        & \textbf{98.0} & 87.6 & 92.5
        & 57.8 & \textbf{92.3} & 71.1
        & 36.7 & 62.4 & 46.2
        & 30.2 & \textbf{82.3} & 44.2
        & 55.7 & \textbf{81.2} & 63.5 \\

        \midrule

        PatternRM, thinking
        & Text only
        & 96.8 & 88.2 & 92.1
        & \textbf{89.5} & 78.1 & 83.4
        & \underline{55.0} & 61.5 & \underline{58.1}
        & \underline{59.0} & 40.5 & \underline{48.0}
        & \underline{75.1} & 67.1 & 70.2 \\

        \rowcolor{gray!15}
        PatternRM, direct
        & Text only
        & 97.6 & \textbf{89.2} & \textbf{93.3}
        & \underline{88.0} & 80.2 & \underline{83.9}
        & \textbf{60.0} & 51.0 & 55.1
        & 58.0 & 48.5 & \textbf{52.8}
        & \textbf{75.9} & 67.2 & \textbf{71.3} \\

        \bottomrule
    \end{tabular}
    \end{adjustbox}
\end{table}

\subsection{PatternRL}

\paragraph{Reward design.}
The training reward consists of a verifier reward for answer correctness and an auxiliary pattern reward for response quality. The verifier extracts the final answer from the last \verb|\boxed{}| expression and applies normalized answer matching and symbolic equivalence checking, while unresolved cases are evaluated by GPT-oss-120B, producing a binary score $s_{\mathrm{ver}}\in\{0,1\}$. For the four PatternRM predictions $\hat b^{\mathrm{RM}}_p$, we use
\begin{equation}
\label{eq:patternrl-weights}
w_{\mathrm{CoT}}=w_{\mathrm{Rep}}=0.05,
\qquad
w_{\mathrm{Con}}=w_{\mathrm{PR}}=0.02.
\end{equation}
To reduce evaluation cost, let $z\sim\operatorname{Bernoulli}(0.6)$ denote whether PatternRM is invoked for a rollout. The auxiliary reward is
\begin{equation}
\label{eq:patternrm-reward}
s_{\mathrm{pat}}
=
-z\min\!\left(0.1,\sum_{p}w_p \hat b^{\mathrm{RM}}_p\right)
\in[-0.1,0].
\end{equation}
The final reward is
\begin{equation}
\label{eq:patternrl-reward}
r=\operatorname{clip}\left(s_{\mathrm{ver}}+s_{\mathrm{pat}},~~0,~~1\right).
\end{equation}
Therefore, incorrect responses always receive zero reward, whereas correct responses receive a score between $0.9$ and $1$, allowing the auxiliary reward to penalize undesirable response patterns without overriding the primary correctness objective.

\paragraph{Training details.}
We initialize the policies from the Qwen3-VL-4B-Instruct and Qwen3-VL-8B-Instruct models and train them with GRPO~\citep{grpo} on a multimodal RL mixture of 44,200 prompts, approximately balanced across multimodal math, multimodal logic, and document understanding. Most samples are drawn from OpenMMReasoner-RL-74K~\citep{zhang2025openmmreasoner}, excluding its \texttt{virl39k} subset, and are supplemented with data from WeMath, MMK12, ThinkLite-VL-Hard-11K, PuzzleVQA, AlgoPuzzleVQA, TextbookQA, ChartQA, and InfographicVQA~\citep{qiao2024wemath,meng2025mmeureka,wang2025thinklite,chia2024puzzlevqa,ghosal2024algopuzzlevqa,kembhavi2017tqa,masry2022chartqa,mathew2022infographicvqa}. The 4B and 8B models are trained for 600 and 400 steps, respectively. We define \textbf{BaseRL} as the correctness-only GRPO baseline trained with the same setup and verifier reward but without $s_{\mathrm{pat}}$. The reward configuration and available training parameters are provided in \Cref{app:patternrl-reward-configuration,app:patternrl-parameters}; the complete 8B configuration is unavailable.

\subsection{PatternRL Results}

\begin{table}[H]
    \centering
    \scriptsize
    \caption{
    \textbf{PatternRL results on PatternEval.}
    All values are percentages; gaps are percentage points. For each backbone, the thinking result is shown as a fixed reference for the non-thinking settings; thinking-mode results after reinforcement learning are not reported.
    }

    \vspace{1mm}
    \label{tab:patternrl_summary}
    \begin{adjustbox}{width=0.95\textwidth}
    \begin{tabular}{l l c cc cccc ccc}
        \toprule
        \multirow{2}{*}{\textbf{Model}}
        & \multirow{2}{*}{\textbf{Setting}}
        & \multirow{2}{*}{\shortstack{\textbf{Gap to}\\\textbf{T ref.}}}
        & \multicolumn{2}{c}{\textbf{Overall Results}}
        & \multicolumn{4}{c}{\textbf{Pattern Failures}}
        & \multicolumn{3}{c}{\textbf{Task-family Trigger}} \\
        \cmidrule(lr){4-5}
        \cmidrule(lr){6-9}
        \cmidrule(lr){10-12}
        & & & Acc$\uparrow$ & Trigger$\downarrow$
        & CoT$\downarrow$ & Rep$\downarrow$ & Con$\downarrow$ & PR$\downarrow$
        & VG$\downarrow$ & OS$\downarrow$ & KR$\downarrow$ \\
        \midrule

        \multirow{4}{*}{Qwen3-VL-4B}
        & Thinking & --
        & 45.20 & \textbf{21.78}
        & 9.98 & 11.68 & 5.38 & 13.62
        & 26.95 & 6.72 & 25.80 \\
        & Non-thinking & 22.44
        & 45.71 & 44.22
        & 25.80 & 33.00 & 12.96 & 20.12
        & 49.26 & 20.35 & 56.18 \\
        & \quad + BaseRL & 29.77
        & \textbf{48.05} & 51.55
        & 33.83 & 43.77 & 13.83 & 20.99
        & 56.37 & 27.59 & 63.97 \\
        & \quad + PatternRL & \textbf{16.69}
        & 47.30 & 38.47
        & 24.51 & 18.84 & 11.01 & 15.61
        & 42.31 & 17.41 & 50.08 \\
        \midrule

        \multirow{4}{*}{Qwen3-VL-8B}
        & Thinking & --
        & 47.55 & \textbf{15.98}
        & 6.21 & 8.16 & 4.06 & 10.06
        & 21.86 & 4.14 & 15.88 \\
        & Non-thinking & 20.79
        & 49.37 & 36.77
        & 23.44 & 25.34 & 9.65 & 15.90
        & 41.13 & 15.52 & 47.63 \\
        & \quad + BaseRL& 28.35
        & 50.57 & 44.33
        & 31.50 & 36.27 & 9.91 & 17.25
        & 49.46 & 17.41 & 56.95 \\
        & \quad + PatternRL & \textbf{14.00}
        & \textbf{50.71} & 29.98
        & 17.23 & 15.24 & 10.85 & 12.05
        & 35.53 & 10.69 & 36.95 \\
        \bottomrule
    \end{tabular}
    \end{adjustbox}
\end{table}

\paragraph{Results on PatternEval.}
As shown in \Cref{tab:patternrl_summary}, correctness-only BaseRL improves task accuracy but aggravates non-thinking response-pattern failures on both Qwen3-VL backbones. In contrast, PatternRL substantially reduces the overall Trigger rate while preserving comparable accuracy, with improvements spanning most failure types and task families. These results show that explicitly optimizing response patterns can mitigate the degradation in quality introduced by correctness-only reinforcement learning without compromising task performance.

\begin{table*}[htbp]
    \centering
    \scriptsize
    \caption{
        \textbf{Accuracy across ten reasoning and document-understanding benchmarks.}
        Bold and underlined values denote the best and second-best results within each model scale, respectively.
    }
    \label{tab:patternrl_task_accuracy}
    \vspace{1mm}

    \setlength{\tabcolsep}{2.5pt}
    \renewcommand{\arraystretch}{1.05}

    \begin{adjustbox}{max width=\textwidth,center}
    \begin{tabular}{l ccccc ccc ccccc c}
        \toprule
        \multirow{2}{*}{\textbf{Method}}
        & \multicolumn{5}{c}{\textbf{Math Reasoning}}
        & \multicolumn{3}{c}{\textbf{Logic Reasoning}}
        & \multicolumn{5}{c}{\textbf{Document Understanding}}
        & \multirow{2}{*}{\shortstack{\textbf{Overall}\\\textbf{Acc.}}}
        \\
        \cmidrule(lr){2-6}
        \cmidrule(lr){7-9}
        \cmidrule(lr){10-14}

        & MathVista
        & MathVerse
        & DynaMath
        & WeMath
        & \textbf{Avg.}
        & LogicVista
        & VisuLogic
        & \textbf{Avg.}
        & AI2D
        & ChartQA
        & DocVQA
        & InfoVQA
        & \textbf{Avg.}
        &
        \\
        \midrule

        \multicolumn{15}{c}{\emph{\textbf{Qwen3-VL-4B}}}
        \\
        Non-thinking
        & 82.28
        & 69.91
        & 73.93
        & \underline{88.35}
        & 78.62
        & 81.32
        & \textbf{28.29}
        & 54.80
        & 89.19
        & 80.30
        & \textbf{84.12}
        & \underline{80.56}
        & \textbf{83.54}
        & 75.83
        \\
        \quad + BaseRL
        & \textbf{86.35}
        & \textbf{73.08}
        & \textbf{79.73}
        & \textbf{91.99}
        & \textbf{82.79}
        & \textbf{89.56}
        & \underline{25.69}
        & \textbf{57.62}
        & \textbf{90.10}
        & \underline{84.73}
        & \underline{75.96}
        & \textbf{81.72}
        & \underline{83.13}
        & \textbf{77.89}
        \\
        \quad + PatternRL
        & \underline{85.56}
        & \underline{70.21}
        & \underline{77.54}
        & 86.06
        & \underline{79.84}
        & \underline{88.46}
        & 23.70
        & \underline{56.08}
        & \underline{89.80}
        & \textbf{85.42}
        & 74.60
        & 78.86
        & 82.17
        & \underline{76.02}
        \\

        \midrule
        \multicolumn{15}{c}{\emph{\textbf{Qwen3-VL-8B}}}
        \\
        Non-thinking
        & 83.07
        & 62.61
        & 75.66
        & 89.49
        & 77.71
        & 85.71
        & \textbf{26.61}
        & \underline{56.16}
        & 87.85
        & 78.45
        & \underline{91.90}
        & 85.75
        & 85.99
        & 76.71
        \\
        \quad + BaseRL
        & \underline{85.17}
        & \textbf{72.09}
        & \textbf{79.98}
        & \textbf{91.57}
        & \textbf{82.20}
        & \textbf{89.01}
        & 25.23
        & \textbf{57.12}
        & \textbf{89.76}
        & \textbf{84.34}
        & 91.23
        & \underline{87.49}
        & \underline{88.20}
        & \textbf{79.59}
        \\
        \quad + PatternRL
        & \textbf{86.61}
        & \underline{68.33}
        & \underline{77.85}
        & \underline{90.63}
        & \underline{80.86}
        & \underline{86.26}
        & \underline{25.38}
        & 55.82
        & \underline{88.98}
        & \underline{84.30}
        & \textbf{92.54}
        & \textbf{88.05}
        & \textbf{88.47}
        & \underline{78.89}
        \\

        \bottomrule
    \end{tabular}
    \end{adjustbox}
\end{table*}

\paragraph{Accuracy on general reasoning tasks.}
As shown in \Cref{tab:patternrl_task_accuracy}, incorporating the response-pattern objective introduces an accuracy trade-off relative to correctness-only BaseRL. The effect is more pronounced for Qwen3-VL-4B, where PatternRL yields lower average accuracy across all three task categories. In contrast, the 8B model largely preserves its task performance, with only limited declines in math and logic reasoning and a slight improvement in document understanding. Overall, although the degradation is substantially smaller at the larger model scale, PatternRL still weakens aggregate task accuracy.

\subsection{Discussion}

\paragraph{Limitations of RL-stage pattern alignment.}
Although PatternRL consistently reduces response-pattern failures, it does not eliminate them entirely, as shown in \Cref{tab:patternrl_summary}. This residual gap suggests that undesirable response patterns cannot be fully corrected through a lightweight auxiliary reward applied only during reinforcement learning. Such behaviors may already be embedded in the model through noisy, inconsistent, or pattern-biased data encountered during earlier training stages. Consequently, more fundamental improvements may require curating higher-quality supervision and incorporating response-pattern constraints during mid-training or supervised fine-tuning, before these behaviors become reinforced by subsequent correctness-oriented optimization.

\paragraph{Capacity dependent correctness--pattern trade-off.}
The cost of introducing pattern-aware optimization also depends on model capacity. As shown in \Cref{tab:patternrl_task_accuracy}, the 4B model exhibits a substantially larger accuracy decline than the 8B model, particularly on math and other reasoning-intensive tasks. One plausible explanation is that PatternRM restricts part of the non-thinking policy's effective rollout space: exploratory, verbose, or partially self-correcting trajectories may help a smaller model reach the correct answer, while also being more likely to trigger response-pattern penalties. Larger models can more readily produce concise and well-structured solutions without relying on these behaviors and therefore better tolerate the additional constraint. PatternRL thus introduces an inherent trade-off between the verifier signal, which rewards successful problem solving, and the PatternRM signal, which constrains how that solution is expressed; balancing the two objectives may need to be adjusted according to model capacity and task difficulty.

\FloatBarrier


\section{Conclusion}
\label{sec:conclusion}

We study response-pattern alignment in hybrid-thinking MLLMs and introduce \textbf{PatternEval} to evaluate four user-visible failures across matched thinking and non-thinking interfaces. Our results reveal a consistent mode-dependent gap: non-thinking inference produces substantially more response-pattern failures, even in frontier models. To mitigate this issue, we develop \textbf{PatternRM} and \textbf{PatternRL}, which reduce non-thinking failures while largely preserving task accuracy. Overall, our findings show that controllable reasoning effort should not come at the cost of unstable response behavior, and that explicitly optimizing response patterns provides a practical path toward more reliable hybrid-thinking MLLMs.

\newpage
\bibliographystyle{plainnat}
\bibliography{reference}

\newpage
\appendix

\section{Appendix Roadmap}
\label{app:roadmap}

The appendix is organized as follows. \Cref{app:more-results} provides the constituent-category PatternEval results for all 25 model configurations, and \Cref{app:meta-judge-analysis} reports the calibration analysis used to select the operational pattern judge. \Cref{app:patternrl-training} documents the PatternRM supervision corpus and the available PatternRL training configuration. Finally, \Cref{app:meta-judge-prompt} reproduces the complete operational pattern-judge prompt in English and in the original Chinese. This roadmap is intended to make the supplementary material navigable without introducing a separate table of contents.

\section{Additional Results}
\label{app:more-results}

\subsection{Category-wise Results}

As shown in \Cref{tab:classwise_pattern_results}, positive cross-mode Trigger gaps occur across multiple constituent task categories rather than only one task family. Some of the largest descriptive differences appear in OOD perception and knowledge-intensive reasoning, whereas content recognition and chart understanding often show smaller gaps. These category-level point estimates characterize PatternEval and do not by themselves identify the cause of a mode difference.

\Cref{tab:classwise_pattern_results} also shows that scale does not uniformly reduce every constituent-category Trigger value. Within several model families, thinking-mode Trigger declines with scale, while non-thinking values remain elevated or vary non-monotonically in categories such as OOD perception, STEM, and general reasoning. This descriptive pattern motivates category-specific analysis rather than a single aggregate interpretation.

\begin{table}[htb]
    \centering
    \scriptsize
    \caption{
        \textbf{Constituent-category Trigger under thinking and non-thinking
        inference.} Values are percentages; lower is better. Categories are
        grouped into the three task families of PatternEval
        (Table~\ref{tab:task_family_composition}).
    }
    \vspace{1mm}
    \label{tab:classwise_pattern_results}
    \setlength{\tabcolsep}{3.0pt}
    \renewcommand{\arraystretch}{0.98}
    \begin{adjustbox}{max width=\textwidth}
    \begin{tabular}{l c cccc cc ccc}
        \toprule
        \multirow{2}{*}{\textbf{Model}} & \multirow{2}{*}{\textbf{Mode}}
        & \multicolumn{4}{c}{\textbf{VG} }
        & \multicolumn{2}{c}{\textbf{OS} }
        & \multicolumn{3}{c}{\textbf{KR} } \\
        \cmidrule(lr){3-6}
        \cmidrule(lr){7-8}
        \cmidrule(lr){9-11}
        & & OOD$\downarrow$ & Rec.$\downarrow$ & Grd.$\downarrow$ & Fact.$\downarrow$
          & OCR$\downarrow$ & Chart$\downarrow$
          & STEM$\downarrow$ & Know.$\downarrow$ & Reas.$\downarrow$ \\
        \midrule
        \multirow{2}{*}{Qwen3.5-0.8B}
        & \NTicon & 98.10 & 81.14 & 63.20 & 69.09 & 58.33 & 41.79 & 94.50 & 79.46 & 78.15 \\
        & \Ticon & 86.86 & 74.86 & 52.00 & 69.09 & 58.33 & 48.57 & 91.50 & 74.05 & 55.19 \\
        \addlinespace[0.5pt]
        \multirow{2}{*}{Qwen3.5-4B}
        & \NTicon & 83.05 & 28.86 & 21.20 & 14.55 & 17.00 & 21.79 & 77.50 & 43.78 & 64.81 \\
        & \Ticon & 44.38 & 16.00 & 4.80 & 0.00 & 6.33 & 10.71 & 25.50 & 25.41 & 32.59 \\
        \addlinespace[0.5pt]
        \multirow{2}{*}{Qwen3.5-9B}
        & \NTicon & 72.76 & 12.86 & 14.00 & 21.82 & 16.67 & 10.00 & 59.50 & 32.43 & 52.96 \\
        & \Ticon & 27.81 & 11.14 & 4.80 & 5.45 & 2.67 & 6.07 & 21.50 & 13.04 & 24.44 \\
        \addlinespace[0.5pt]
        \multirow{2}{*}{Qwen3.5-27B}
        & \NTicon & 54.29 & 6.86 & 18.80 & 16.36 & 10.00 & 4.64 & 65.50 & 16.22 & 51.48 \\
        & \Ticon & 9.14 & 2.29 & 2.80 & 3.64 & 1.67 & 0.72 & 8.50 & 1.63 & 13.38 \\
        \addlinespace[0.5pt]
        \multirow{2}{*}{Qwen3.5-35B-A3B}
        & \NTicon & 65.14 & 6.86 & 20.00 & 16.36 & 13.67 & 11.43 & 60.00 & 17.84 & 54.07 \\
        & \Ticon & 15.25 & 2.29 & 2.00 & 1.82 & 1.67 & 1.43 & 8.50 & 3.78 & 10.23 \\
        \addlinespace[0.5pt]
        \multirow{2}{*}{Qwen3.5-122B-A10B}
        & \NTicon & 55.62 & 5.43 & 16.40 & 21.82 & 10.00 & 7.14 & 58.50 & 15.14 & 47.41 \\
        & \Ticon & 9.63 & 1.43 & 2.00 & 1.82 & 0.67 & 1.07 & 11.50 & 4.35 & 16.67 \\
        \addlinespace[0.5pt]
        \multirow{2}{*}{Qwen3.5-397B-A17B}
        & \NTicon & 58.67 & 2.00 & 15.60 & 10.91 & 11.67 & 7.14 & 61.50 & 9.19 & 56.67 \\
        & \Ticon & 8.57 & 1.43 & 2.40 & 0.00 & 1.00 & 0.36 & 4.00 & 2.70 & 10.78 \\
        \midrule
        \multirow{2}{*}{Qwen3.6-27B}
        & \NTicon & 65.14 & 8.31 & 16.00 & 14.55 & 13.33 & 7.86 & 63.00 & 16.76 & 57.04 \\
        & \Ticon & 12.26 & 2.00 & 2.80 & 3.64 & 1.67 & 1.43 & 8.00 & 3.24 & 11.85 \\
        \addlinespace[0.5pt]
        \multirow{2}{*}{Qwen3.6-35B-A3B}
        & \NTicon & 69.52 & 5.43 & 17.20 & 20.00 & 12.00 & 8.57 & 59.00 & 14.05 & 55.56 \\
        & \Ticon & 10.56 & 2.29 & 3.20 & 1.82 & 0.67 & 1.79 & 10.50 & 5.41 & 12.12 \\
        \addlinespace[0.5pt]
        \multirow{2}{*}{Qwen3.7-Max}
        & \NTicon & 50.29 & 6.59 & 18.80 & 29.09 & 13.33 & 13.19 & 29.50 & 8.15 & 53.33 \\
        & \Ticon & 9.94 & 1.43 & 7.20 & 3.64 & 3.00 & 1.43 & 6.50 & 4.35 & 8.52 \\
        \addlinespace[0.5pt]
        \multirow{2}{*}{Qwen3.7-Plus}
        & \NTicon & 52.19 & 3.71 & 15.60 & 34.55 & 14.33 & 13.62 & 31.50 & 15.22 & 57.25 \\
        & \Ticon & 14.48 & 0.86 & 3.60 & 3.64 & 1.33 & 2.15 & 10.50 & 3.26 & 11.90 \\
        \midrule
        \multirow{2}{*}{Qwen3-VL-30B-A3B}
        & \NTicon & 69.20 & 12.29 & 31.60 & 52.73 & 30.33 & 18.57 & 76.65 & 25.95 & 60.67 \\
        & \Ticon & 30.19 & 4.00 & 1.60 & 5.45 & 4.67 & 1.07 & 9.00 & 10.81 & 8.89 \\
        \addlinespace[0.5pt]
        \multirow{2}{*}{Qwen3-VL-32B}
        & \NTicon & 70.75 & 10.57 & 24.40 & 29.09 & 15.00 & 10.36 & 59.00 & 16.85 & 51.11 \\
        & \Ticon & 12.83 & 2.86 & 0.00 & 7.27 & 0.67 & 2.14 & 3.89 & 8.11 & 4.85 \\
        \midrule
        \multirow{2}{*}{Kimi-K2.5}
        & \NTicon & 82.86 & 2.86 & 23.20 & 34.55 & 35.67 & 11.79 & 82.00 & 27.03 & 72.96 \\
        & \Ticon & 11.89 & 1.72 & 4.00 & 1.85 & 4.00 & 1.43 & 4.10 & 1.63 & 15.67 \\
        \addlinespace[0.5pt]
        \multirow{2}{*}{Kimi-K2.6}
        & \NTicon & 83.05 & 3.15 & 30.80 & 74.55 & 37.67 & 18.57 & 85.50 & 32.43 & 87.78 \\
        & \Ticon & 2.44 & 0.29 & 1.20 & 1.82 & 0.67 & 0.00 & 0.00 & 1.08 & 1.47 \\
        \midrule
        \multirow{2}{*}{Seed-1.6-Vision}
        & \NTicon & 76.57 & 6.63 & 20.00 & 21.82 & 11.67 & 9.29 & 33.50 & 15.76 & 25.79 \\
        & \Ticon & 21.33 & 6.34 & 10.80 & 9.09 & 2.67 & 3.57 & 7.50 & 7.07 & 3.97 \\
        \addlinespace[0.5pt]
        \multirow{2}{*}{Seed-2.0-Pro}
        & \NTicon & 10.86 & 1.15 & 8.80 & 5.45 & 2.67 & 4.64 & 10.50 & 8.15 & 9.13 \\
        & \Ticon & 3.62 & 1.73 & 8.40 & 0.00 & 1.67 & 0.00 & 5.50 & 1.63 & 3.97 \\
        \addlinespace[0.5pt]
        \multirow{2}{*}{Seed-2.1-Pro}
        & \NTicon & 19.62 & 3.17 & 13.60 & 5.45 & 2.67 & 4.29 & 19.50 & 10.33 & 17.06 \\
        & \Ticon & 1.39 & 0.29 & 5.65 & 1.89 & 3.33 & 1.07 & 4.52 & 1.66 & 1.80 \\
        \midrule
        \multirow{2}{*}{Claude-Opus-4.6}
        & \NTicon & 75.92 & 6.12 & 15.32 & 12.73 & 7.41 & 3.21 & 59.80 & 11.24 & 62.75 \\
        & \Ticon & 8.62 & 1.17 & 2.82 & 12.73 & 1.35 & 1.07 & 0.56 & 1.70 & 8.26 \\
        \addlinespace[0.5pt]
        \multirow{2}{*}{Claude-Opus-4.8}
        & \NTicon & 57.61 & 2.04 & 10.48 & 10.91 & 7.41 & 1.79 & 30.65 & 3.37 & 36.44 \\
        & \Ticon & 7.74 & 0.87 & 2.42 & 5.45 & 0.67 & 2.50 & 1.52 & 0.56 & 8.91 \\
        \midrule
        \multirow{2}{*}{Mimo-v2-Omni}
        & \NTicon & 39.85 & 15.19 & 16.06 & 34.55 & 9.67 & 19.06 & 24.00 & 9.19 & 30.86 \\
        & \Ticon & 24.04 & 2.22 & 8.29 & 9.09 & 3.33 & 1.98 & 7.00 & 4.32 & 14.87 \\
        \addlinespace[0.5pt]
        \multirow{2}{*}{Mimo-v2.5}
        & \NTicon & 44.38 & 20.63 & 30.45 & 40.00 & 3.67 & 14.03 & 31.50 & 11.89 & 23.42 \\
        & \Ticon & 30.10 & 3.15 & 3.79 & 3.64 & 2.67 & 1.87 & 3.50 & 3.91 & 20.31 \\
        \midrule
        \multirow{2}{*}{GPT-5 Nano}
        & \NTicon & 77.90 & 12.64 & 10.00 & 7.27 & 30.77 & 14.64 & 55.00 & 25.41 & 32.71 \\
        & \Ticon & 42.29 & 0.86 & 2.00 & 5.45 & 2.01 & 2.86 & 7.00 & 5.95 & 11.15 \\
        \addlinespace[0.5pt]
        \multirow{2}{*}{GPT-5.4}
        & \NTicon & 9.71 & 3.43 & 0.80 & 3.64 & 1.67 & 0.71 & 12.00 & 6.49 & 7.04 \\
        & \Ticon & 4.95 & 3.71 & 0.80 & 1.82 & 1.33 & 0.36 & 3.00 & 1.08 & 1.85 \\
        \addlinespace[0.5pt]
        \multirow{2}{*}{GPT-5.5}
        & \NTicon & 14.86 & 1.74 & 0.40 & 1.82 & 1.34 & 1.07 & 4.50 & 3.83 & 4.49 \\
        & \Ticon & 1.61 & 1.16 & 0.40 & 1.82 & 1.01 & 0.00 & 0.51 & 0.55 & 0.00 \\
        \bottomrule
    \end{tabular}
    \end{adjustbox}
\end{table}

\section{Meta-Judge Analysis}
\label{app:meta-judge-analysis}

\subsection{Calibration Protocol}
\label{app:meta-judge-calibration-protocol}
The calibration pipeline separates candidate discovery from benchmark scoring. During candidate discovery, Seed-2.0-Pro, Kimi-K2.6, and Qwen3.5-397B independently annotate responses along the four PatternEval failure labels. Complete three-judge agreement defines the consensus stratum, while any disagreement defines the disputed stratum; the two strata are sampled at an approximate $7{:}3$ ratio, with the predicted positive rate controlled near $70\%$. The selected 2,500 fixed responses then receive human reference annotations under the same rubric. After reference labeling, we compare five candidate judge systems---with image and text-only variants where available---on the same responses using per-label precision, recall, and F1. This fixed-response design supports controlled judge comparison, while the reference set remains separate from the 25-configuration benchmark evaluation.

\subsection{Failure-Label Calibration Behavior}
\label{app:meta-judge-category-analysis}
The results in \Cref{tab:meta_judge_candidate_scores} show a consistent difficulty gap across the nine judge--input configurations. CoT leakage F1 ranges from 90.2\% to 95.3\%, and repetition F1 ranges from 81.6\% to 88.2\%. In contrast, contradiction F1 ranges from 56.1\% to 62.5\%, and performative reasoning F1 ranges from 41.8\% to 64.5\%. Image access improves the reported aggregate F1 for Seed-2.0-Pro and Kimi-K2.6, while GPT-5.5 achieves the highest aggregate F1 overall. Seed-2.0-Pro with image access attains the highest contradiction F1 (62.5\%) and directly inspects visual evidence; we therefore use it as the operational pattern judge.

\section{PatternRL Training}
\label{app:patternrl-training}

\subsection{Supervision Corpus}
\label{app:patternrl-supervision-corpus}
PatternRM is initialized from Qwen3.5-27B. Its initial corpus draws 15K examples from each of four completed rollout collections. Within each collection, trajectories are partitioned into three step-count buckets and sampled at a $2{:}3{:}5$ ratio with question-level deduplication. Quality filtering excludes the NED-only subset, retains half of the overlong cases, removes responses shorter than five characters, and removes entirely Chinese responses to English prompts. Prompts associated with filtered responses are then re-rolled out with Qwen3-VL-32B-Instruct, Qwen3.5-4B-Think, and Kimi-K2.6-NoThink, contributing 10K examples per policy. The resulting pool contains 90K responses.

Kimi-K2.6, Seed-2.0-Pro, and Qwen3.5-397B independently assign the four PatternEval labels, and only unanimous four-label vectors are retained. For the retained targets, thinking-format examples use Kimi-K2.6's thinking content as the rationale field, whereas non-thinking-format examples use the reasoning in its answer field. Consensus filtering yields 52,344 unique examples. Reusing 5,234 instances produces 57,578 SFT instances in total, comprising 34,547 thinking-format and 23,031 non-thinking-format targets. The incomplete A20B PatternRM-fusion trajectory collection is excluded from these counts.

\subsection{Reward Configuration}
\label{app:patternrl-reward-configuration}
PatternRL converts the four PatternRM decisions into the category-weighted auxiliary reward defined in \Cref{eq:patternrl-weights,eq:patternrm-reward}. Logical contradiction and performative reasoning each carry weight 0.02, while CoT leakage and repetition each carry weight 0.05. PatternRM is invoked independently with probability $0.6$ for each rollout; otherwise its auxiliary contribution is zero. Simultaneous penalties accumulate until the PatternRM subreward reaches its floor of $-0.1$. The subreward is added to the outcome reward and the fused scalar is clipped to $[0,1]$ as specified in \Cref{eq:patternrl-reward}; consequently, $-0.1$ is the auxiliary-reward floor, whereas zero is the final fused-reward floor. The configuration reported below corresponds to a single run; multi-seed training manifests are unavailable.

\subsection{Training Parameters}
\label{app:patternrl-parameters}
The training configuration used for the reported PatternRL run is summarized in \Cref{tab:patternrl-training-parameters}.

\begin{table}[H]
    \centering
\caption{
    \textbf{PatternRL training configuration.}
    Summary of the optimization hyperparameters, rollout settings, sequence-length budgets, reward configuration, and distributed training setup used for PatternRL.
}    \label{tab:patternrl-training-parameters}
    \begingroup
    \small
    \setlength{\tabcolsep}{4pt}
    \renewcommand{\arraystretch}{0.95}
    \begin{tabular}{@{}ll@{}}
        \toprule
        Setting & Value \\
        \midrule
        Optimization algorithm & GRPO \\
        Optimizer & Adam \\
        Learning rate & $1\times10^{-6}$ \\
        Clip ratio & 0.2 \\
        KL coefficient & 0 \\
        Entropy coefficient & 0 \\
        Training batch size & 128 \\
        PPO mini-batch size & 128 \\
        Micro-batch size per GPU & 1 \\
        Max prompt length & 2,048 \\
        Max response length & 16,384 \\
        Rollouts per prompt & 8 \\
        Temperature & 1.0 \\
        Top-$p$ & 1.0 \\
        Training epochs & 5 \\
        Training steps & 600 \\
        Actor parallelism & $\mathrm{TP}=2$, $\mathrm{PP}=1$ \\
        Rollout parallelism & $\mathrm{TP}=2$ \\
        Random seed & 42 \\
        \bottomrule
    \end{tabular}
    \endgroup
\end{table}

\clearpage
\section{PatternEval Judge Prompt}
\label{app:patterneval-judge-prompt}
\label{app:meta-judge-prompt}

The meta-judge receives the user question, any associated image, and the model response, and independently assigns the four response-pattern labels defined in \Cref{sec:response-pattern-taxonomy}. The image is not serialized into the text placeholders reproduced below; it is supplied separately as the multimodal image input associated with the same request. The complete textual prompt is reproduced below for transparency. We provide a faithful English translation first, followed by the original Chinese prompt used in our evaluation. Formatting has been adapted for typesetting, but the decision rules, priority policy, examples, required output order, and input placeholders are unchanged. Exact API serialization, image preprocessing, and model-version metadata remain part of the release manifest required for independent reproduction.

\subsection{English Version}
\label{app:meta-judge-prompt-en}

\begin{AIbox}{Response-Pattern Meta-Judge Prompt (English Translation)}
\sloppy
You are a strict analyzer of undesirable patterns in model responses. Given the user question, any relevant image, and the model response, determine whether the response exhibits each of the following four undesirable patterns. For every pattern, output a binary decision (1 or 0) and a one-sentence justification.

\medskip
\textbf{I. Decision Principles}
\begin{enumerate}[leftmargin=*,nosep]
    \item Base the judgment only on the question, any relevant image, and the model response. Every decision must be grounded in observable text in the model response.
    \item Use the image to verify image-related descriptions, counts, localization, recognition, and spatial relations in the response. However, do not mark a pattern merely because the answer is incorrect.
    \item Factual errors, visual-recognition errors, calculation errors, and incomplete answers are not undesirable response patterns by themselves. Mark 1 only when the error manifests as one of the named patterns below.
    \item Judge the four labels independently; multiple labels may be positive. Mark 1 only when there is explicit textual evidence; otherwise mark 0.
    \item Length, many steps, or detailed explanation alone do not constitute an undesirable pattern. Do not mark 1 merely because a response is long or contains extensive reasoning.
\end{enumerate}

\medskip
\textbf{II. Decision Order (Important)}

Apply the following fixed order:
\begin{enumerate}[leftmargin=*,nosep]
    \item First judge \textbf{chain-of-thought leakage}: read the full response and check whether it exposes internal deliberation that should remain hidden, including self-correction, tentative wavering, or process-oriented narration that lays out the solution process from the beginning, according to Category~1.
    \item Next judge \textbf{response repetition} independently of whether leakage is present, according to Category~2.
    \item Finally judge \textbf{logical contradiction} and \textbf{performative reasoning}. First determine whether content that appears contradictory or superficially reasoned is actually part of exposed internal deliberation, such as self-correction, repeated trial, process-level wavering, or draft-like reasoning. If so, attribute that content to chain-of-thought leakage and do not count it again as contradiction or performative reasoning; mark those categories 0 for that content. An exception applies when the final delivered answer contains a separate hard contradiction or an independently unsupported shell of reasoning. Logical contradiction concerns only unresolved, mutually incompatible claims retained in the final delivered content. Performative reasoning concerns analysis that adds no valid support for the conclusion, or staged analysis on a low-reasoning task, when chain-of-thought leakage is not the better explanation.
\end{enumerate}

\medskip
\textbf{III. Four Undesirable Patterns}

\medskip
\textbf{1. Chain-of-Thought Leakage}

\textbf{Definition.} The model directly exposes intermediate reasoning, internal self-talk, tentative analysis, task decomposition, self-correction, or draft-like thinking that should remain hidden, thereby reducing the naturalness of the response or violating the expected output style.

\textbf{Mark 1 only when there is explicit evidence of exposed internal reasoning in at least one of the following forms:}
\begin{itemize}[leftmargin=*,nosep]
    \item \textbf{First-person process narration:} phrases such as ``let me look first,'' ``I will analyze this first,'' ``let me think,'' ``let me check,'' ``I need to judge this,'' or ``I need to confirm this first''; or reader-guiding phrases such as ``let us analyze these one by one,'' ``let us analyze step by step,'' ``let us first look at,'' and similar uses of ``let me,'' ``let us,'' or ``we first.'' These expressions narrate how the model is looking or thinking and lead the reader through the model's working process. Once such process narration appears, it counts as leakage even if followed by bullet points or section headings.
    \item \textbf{Self-correction, reversal, wavering, or repeated trial:} traces such as ``no, let me look again,'' ``what I said earlier was wrong; it should be,'' ``wait, let me reconsider,'' or ``try A first; no, then try B.''
    \item \textbf{Exposed internal decisions or drafts:} phrases such as ``end of thinking,'' ``now confirmed,'' ``my preliminary judgment is,'' or ``let me run through it mentally,'' as well as internal decisions, execution strategies, task decomposition, draft reasoning, or residual markers such as \texttt{think}.
\end{itemize}

\textbf{Mark 0 in the following cases, provided that none of the above traces is present:}
\begin{itemize}[leftmargin=*,nosep]
    \item User-facing structured solution or analysis: sections such as ``Analysis,'' ``Step 1,'' and ``Step 2,'' or objective statements such as ``to determine the conclusion, first analyze \ldots,'' are not leakage when the response is a polished explanation directed to the user, without first-person process narration, self-correction, wavering, or exposed internal decisions. If such analysis is unsupported, classify it as performative reasoning rather than leakage.
    \item The user explicitly requests reasoning, steps, a draft, or a method, and the response provides a polished, user-facing derivation, tutorial, procedure, or solution.
    \item A direct conclusion with a brief justification or concise supporting points.
    \item A short elimination of alternatives, comparison, or user-facing explanation of multiple perspectives or conditional branches.
\end{itemize}

\textbf{Positive example.} The question asks for an object's orientation. The response says, ``Let me look at this image first. It seems to face left---no, looking again, it should face right,'' exposing repeated tentative inspection.

\textbf{Negative example.} The question asks for the orientation of a piece of furniture. The response contains an ``Analysis'' section and objectively derives the orientation in Step~1 and Step~2 without first-person self-talk or self-correction. This is not leakage; if the derivation provides no valid image evidence, it is performative reasoning.

\medskip
\textbf{2. Response Repetition}

\textbf{Definition.} The response unnecessarily repeats the same sentence, meaning, structure, or conclusion to a degree that clearly harms information efficiency, readability, or answer quality.

\textbf{Mark 1 in the following cases:}
\begin{itemize}[leftmargin=*,nosep]
    \item The same conclusion, recommendation, or disclaimer is repeated consecutively or frequently.
    \item The same information or conclusion is restated across multiple sections---for example, analysis, verification, summary, and final answer---without adding information, even if the wording changes.
    \item Paragraphs are highly homogeneous and merely paraphrase one another.
    \item The reasoning circles back, repeatedly lists the same candidates or clues, or repeatedly rejects and recomputes the same result beyond what normal emphasis requires, producing a mechanical or cyclic pattern.
\end{itemize}

\textbf{Mark 0 in the following cases:}
\begin{itemize}[leftmargin=*,nosep]
    \item Limited repetition used to emphasize a key point.
    \item Necessary local repetition in an enumeration or stepwise structure when the items differ substantively.
    \item A single useful self-correction or one concluding restatement that naturally connects the reasoning to the final answer without creating obvious redundancy.
\end{itemize}

\textbf{Positive example.} When asked for a painting's name, the response gives the title and artist in a heading, then immediately repeats the same title, artist, and year in a ``Basic Information'' list without adding information.

\textbf{Negative example.} When asked for the answer to a problem, the response provides a derivation and closes with one sentence restating the final answer.

\medskip
\textbf{3. Logical Contradiction}

\textbf{Definition.} In the final delivered content, the response gives two claims about the same object, value, or conclusion that cannot both be true under the same conditions. Both claims remain presented as valid information: the response does not choose between them, invalidate either one, or distinguish their conditions. A positive decision requires locating two specific, mutually incompatible passages, A and B.

\textbf{A hard contradiction must satisfy all of the following conditions:} A and B concern the same object, quantity, or conclusion; they use the same conditions, time, and measurement convention; they cannot both be true; the model does not state that one has been withdrawn or is merely an alternative; and both remain active claims in the answer.

\textbf{Mark 1 for any of the following four forms:}
\begin{enumerate}[leftmargin=*,nosep]
    \item \textbf{Mutually exclusive final conclusions.} The same question, object, or answer position receives two incompatible final conclusions without a choice. For example, the response first identifies an artifact as a bronze \emph{zun} but ends with ``Final answer: bronze \emph{ding},'' or first states that A is correct and B violates the conditions but ultimately selects B.
    \item \textbf{Incompatible attributes of the same object.} The response assigns mutually exclusive values of the same attribute to one object and retains both. For example, it calls block \(b\) horizontal in a Klotski puzzle and later instructs the user to move the vertical block \(b\); or it says a function is odd and symmetric about the origin, then says it is symmetric about the \(y\)-axis and therefore even.
    \item \textbf{Incompatible values of the same quantity.} The response retains two conflicting values for the same number, date, coordinate, count, angle, area, or temperature. For example, it uses two different base areas without a correction; states that an image contains six circles but itemizes three on the left and four on the right for a total of seven; or gives point \(P\) first as \((2,3)\) and later as \((3,2)\).
    \item \textbf{An unretracted derivation conflicts with the final answer.} The reasoning explicitly derives and retains \(X\), but the final answer gives incompatible \(Y\) without abandoning the earlier result. For example, the steps identify 8 and 9 as mandatory waypoints but the final waypoint list omits them; or the derivation obtains \(x=3\), verifies it by substitution, and then gives \(x=5\) as the final answer.
\end{enumerate}

\textbf{Mark 0 in the following cases:}
\begin{itemize}[leftmargin=*,nosep]
    \item \textbf{Corrected, abandoned, or resolved alternatives.} An earlier statement is explicitly rejected, corrected, or abandoned, or one candidate is selected from several and only one final conclusion remains. Process-level wavering of this kind belongs to chain-of-thought leakage.
    \item \textbf{Hedging or uncertainty.} Expressions such as ``possibly,'' ``probably,'' ``more likely,'' ``tends to be,'' or ``cannot rule out'' indicate alternatives or confidence differences rather than two definite incompatible conclusions. For example, ``possibly late Shang, though early Western Zhou cannot be ruled out.''
    \item \textbf{Different conditions, times, viewpoints, or conventions are distinguished.} For example, the result is \(X\) under Assumption~1 and \(Y\) under Assumption~2; or an object is on the left from the observer's viewpoint but on the right relative to its own facing direction.
    \item \textbf{A single wrong answer, visual error, or calculation error without an internal contradiction.} This category does not evaluate correctness. Mark 0 if the answer is wrong but does not retain two mutually incompatible claims about the same object, quantity, or conclusion.
    \item \textbf{Imprecision, equivalent restatement, unit conversion, repetition, rough reasoning, or performative reasoning without a locatable pair of incompatible claims.} Examples include calling an altocumulus cloud a ``mackerel sky'' or stating that one meter equals 100 centimeters.
\end{itemize}

\textbf{Key distinction.} A hard contradiction retains two propositions about the same object attribute, quantity, or conclusion that cannot both be true. Process-level wavering, hedging, visual or factual errors, distinguished conditions, and resolved alternatives do not qualify.

\textbf{Positive example.} When asked which position from the left contains the white-haired person wearing glasses, the response identifies the target as item 4 in its list but concludes ``therefore, the fifth.'' It retains two incompatible answers, 4 and 5, without resolving them.

\textbf{Negative example.} When asked which animal appears in an image, the response consistently answers ``a golden retriever'' and adds supporting details such as long fur and golden coloring.

\medskip
\textbf{4. Performative Reasoning}

\textbf{Definition.} The response presents the appearance of analysis, derivation, argument, or evidential support, but that material adds no valid information that bears on the final conclusion. It performs being well-supported rather than providing support.

\textbf{Valid information gain} means an observation, comparison, count, elimination, disambiguation, calculation, or verification that can support the conclusion. For image tasks, valid support must refer to clearly observable content, such as specific objects, text, colors, quantities, positions, relations, chart values, or paths. It may compare size, distance, occlusion, direction, shape, or state; reason from visible paths, walls, nodes, arrows, table rows or columns, and flow branches; deduplicate or match people and objects across images; or read text, legends, coordinates, labels, and values. Generic phrases such as ``as can be seen from the image,'' ``considering the overall layout,'' ``based on visual characteristics,'' or ``after comprehensive judgment'' usually constitute performative reasoning when they identify no concrete visible content.

\textbf{Mark 1 only when the response adopts an analytical, inferential, argumentative, or evidence-bearing posture and exhibits one of the following forms:}
\begin{enumerate}[leftmargin=*,nosep]
    \item \textbf{Unsupported reasoning wrapper.} The response uses the form of reasoning or evidence without substantive support. Typical signs include ``first,'' ``second,'' ``therefore,'' or ``in summary'' connecting only generic assertions; merely restating the question or conclusion; claiming to reason ``from the image'' without naming any concrete element, position, count, text, color, direction, occlusion, or relation; or invoking frameworks such as semantics, structure, and context without connecting them to specific evidence. For example, when asked to identify a plant, the response says that leaf shape, growth habit, and overall visual characteristics establish the species but identifies none of those visible characteristics.
    \item \textbf{Evidence that does not support the conclusion.} The response offers apparently relevant analysis or observations that do not entail the conclusion, or merely packages an answer that could be read directly from the image or was already supplied by the user. Typical cases include analyzing A but concluding B; providing background that cannot establish the answer; identifying something from generic impressions such as ``looks similar,'' ``the style fits,'' or ``the shape is close'' rather than discriminative evidence; adding analysis to a low-reasoning task without any new observation, comparison, count, elimination, disambiguation, or verification; restating a user-provided route without comparing position, roads, distance, reachability, or obstacles; or calling maze nodes ``key'' or ``mandatory'' without checking wall and path connectivity or excluding alternatives. This also includes identification answers padded with biography, work lists, general stylistic history, typical roles, or other background that does not explain how the image establishes the identity through a face, inscription, signature, emblem, or other discriminative detail.
\end{enumerate}

\textbf{Mark 0 in the following cases:}
\begin{itemize}[leftmargin=*,nosep]
    \item A direct answer or brief explanation that does not stage an analysis. Merely saying ``based on the image'' without naming an image element is treated as a direct answer.
    \item Image analysis that cites concrete visible evidence supporting the conclusion, even if brief, approximate, verbose, or imperfect. Examples include ``the lower-right button reads `Submit,' so it is the submit button'' and comparing the water levels and vessel shapes of two cups. The evidence must be sufficiently discriminative, such as visible text, an inscription or signature, an exact count or position, or a comparable state. Generic terms such as ``similar shape,'' ``matching style,'' ``technological feel,'' ``flame-like core,'' or ``similar colors'' do not qualify as valid evidence.
    \item A genuine image-connectivity analysis that reaches the wrong result because of a visual error. Mark 0 only when the response actually traces whether adjacent cells connect through walls or passages and uses that connectivity to rule out alternatives. Merely naming cells or nodes as ``key'' or ``mandatory,'' or listing their locations without proving connectivity or excluding alternatives, remains performative reasoning.
    \item The user explicitly requests justification, analysis, stepwise reasoning, or evidence, and the analysis is not entirely empty; for example, an elimination argument explaining why option C is selected.
    \item A visual, calculation, or factual error when the cited evidence genuinely bears on the conclusion; for example, counting real objects in the image but missing or miscounting one.
    \item A cautious statement that the image is unclear and the conclusion cannot be confirmed, followed by a tentative answer or refusal to assert one.
\end{itemize}

\textbf{Key distinction.} Direct answers, concrete image evidence, valid reasoning, and chain-of-thought leakage receive 0 for this category. Analysis that only restates the question or conclusion, stacks transitions and generic assertions, or claims image evidence without identifying a concrete visible detail receives 1. Evidence may be mistaken yet still be genuine support; do not mark 1 merely because the conclusion is wrong.

\textbf{Positive example.} When asked which button submits a form, the response says that the interface layout, button semantics, and operation path must be considered together and therefore chooses the lower-right button, without naming its text, color, or icon.

\textbf{Negative example.} When asked how many apples appear in an image, the response says that one is on the left and two are on the plate on the right, for a total of three.

\medskip
\textbf{IV. Required Output}

Follow the exact order and format below. Do not output unrelated content and do not omit any category. For each category, first output 1 or 0, then give a one-sentence justification. When a category is positive, identify the key evidence; for logical contradiction, identify both mutually incompatible passages A and B.

\medskip
{\ttfamily
1. Chain-of-thought leakage: 1 / 0\\
Reason:\\[2pt]
2. Response repetition: 1 / 0\\
Reason:\\[2pt]
3. Logical contradiction: 1 / 0\\
Reason:\\[2pt]
4. Performative reasoning: 1 / 0\\
Reason:
}

\medskip
\textbf{V. Input}

\textbf{Question:}\\
\texttt{\{question\}}

\medskip
\textbf{Model response:}\\
\texttt{\{response\}}
\end{AIbox}

\clearpage
\subsection{Chinese Version }
\label{app:meta-judge-prompt-zh}

\begin{AIbox}{Response-Pattern Meta-Judge Prompt (Chinese Original)}
\begin{CJK*}{UTF8}{gbsn}
\renewcommand{\bfdefault}{bx}
\sloppy
你是一个严格的模型回复坏模式分析器。请结合用户问题、相关图片（若有）和模型回答，判断模型回答是否存在以下四类坏模式，并为每一类给出 1 或 0 的判定和一句话理由。

\medskip
\textbf{一、判定原则}
\begin{enumerate}[leftmargin=*,nosep]
    \item 只依据问题、相关图片（若有）和模型回答来判断，判定要落在模型回答的可观察文本上。
    \item 要结合图片核验模型回答里和图像有关的描述、计数、定位、识别、空间关系是否成立；但不要因为答得对不对本身就判坏模式。
    \item 事实错误、看图错误、算错、答得不全，这些本身不算坏模式，只有当它们表现为下面某一类可命名的坏模式时才判 1。
    \item 四个标签各自独立判断，允许同时成立；只有存在明确文本证据时才判 1，否则判 0。
    \item 回答篇幅长、步骤多、解释详细，本身都不是坏模式，不能只因为篇幅长或推理多就判 1。
\end{enumerate}

\medskip
\textbf{二、判定顺序（重要）}

请按下面的固定顺序判断：
\begin{enumerate}[leftmargin=*,nosep]
    \item 先判思维链外溢：通读回答，看是否暴露了本该隐藏的内部思考、自我纠错、试探摇摆，或把解题过程从头摊开的过程腔（按第 1 类标准）。
    \item 再判病态重复：独立判断，和是否外溢无关（按第 2 类标准）。
    \item 再判逻辑性冲突和伪装推理：先看那些看起来像矛盾或像空泛推理的内容，是否其实来自暴露出来的内部思考过程（自我纠错、反复试探、过程摇摆、草稿式推演）。若是，则这部分归思维链外溢、不再计入逻辑性冲突或伪装推理（这两项就该内容判 0），除非回答在最终交付内容里另有独立的硬矛盾或独立的空壳推理。逻辑性冲突只针对最终交付内容里成对并列、无法同时成立、且未取舍的硬矛盾；伪装推理只针对推理对结论没有有效信息增量、或在低推理负担任务上表演式分析，且该问题不能被思维链外溢更好地解释时。
\end{enumerate}

\medskip
\textbf{三、四类坏模式}

\medskip
\textbf{1. 思维链外溢}

\textbf{定义：}模型把本该隐藏的中间推理过程、内部自我对话、试探性分析、任务分解、自我纠错或草稿式思考，直接暴露在最终回复里，影响回答的自然度或不符合输出预期。

\textbf{判 1 必须有明确的内部思考外露痕迹，落在下面任一类上：}
\begin{itemize}[leftmargin=*,nosep]
    \item 第一人称过程腔：让我先看看、我先分析一下、让我想想、我来看看、这里我需要判断一下、我得先确认一下，以及我们来逐一分析、我们来逐步分析、我们先看、我们可以先等以让我、我们来、我们先等开头、把怎么看怎么想讲出来、带着读者一步步做题的表述。这类一旦出现，即使后面分点、带小标题，也算外溢。
    \item 自我纠错、推翻、摇摆、反复试探：不对再看、我前面说错了应该是、等一下重新看、先试 A 不对再试 B 等中途否定再改或来回试探的痕迹。
    \item 暴露内部决策或草稿：结束思考、现在确认、我的初步判断是、先在脑子里过一遍等内部决策、执行策略、任务分解的过程，或残留 think 这类思考、推理标签。
\end{itemize}

\textbf{判 0 的情形（没有上面这些痕迹就判 0）：}
\begin{itemize}[leftmargin=*,nosep]
    \item 面向用户的结构化解题或分析：设了分析过程、步骤一步骤二等板块、把推导分点展开，或以要确定某结论需先分析某某这类非第一人称的客观陈述开头，只要整体是面向用户、成品式地讲题或讲分析（解题步骤、逐项说明、逐图说明），没有让我、我们来这类过程腔、没有自我纠错摇摆、没有暴露内部决策或草稿，就不算外溢。这类回答若分析空转无有效依据，应归伪装推理，不归外溢。
    \item 用户明确要求展示推理、步骤、草稿，且回答是面向用户的成品式步骤、推导或方法说明（教程、操作流程、解题步骤）。
    \item 直接交付结论加简短依据或要点。
    \item 简短的候选排除、比较，或面向用户的多角度、条件分支说明。
\end{itemize}

\textbf{示例（判 1）：}问某物体朝向。回答写让我先看看这张图，嗯好像朝左，不对再看一下应该是朝右，把自己来回试探的过程写了出来。

\textbf{示例（判 0）：}问推断某家具朝向。回答设了分析过程板块，按步骤一步骤二客观地把朝向推导讲给用户，没有第一人称自言自语或自我纠错，则不算外溢（若推导无有效据图依据，归伪装推理）。

\medskip
\textbf{2. 病态重复}

\textbf{定义：}模型在回复里不必要地重复同一句话、同一语义、同一结构或同一结论，这种重复已经明显影响信息效率、阅读体验或回答质量。

\textbf{判 1 的情形：}
\begin{itemize}[leftmargin=*,nosep]
    \item 同一结论、同一建议、同一免责声明被连续或高频重复，反复表达同一个意思。
    \item 同一个信息或结论在多个段落（比如分析、验证、总结、最终答案）反复重申，即使分属不同小节、措辞略有变化，只要没有新增信息，也算重复。
    \item 段落之间内容高度同质，只是换少量措辞重新说一遍。
    \item 推理兜圈、候选和线索反复罗列、同一计算反复否定重来等，超出正常强调需要，呈机械化、循环化倾向。
\end{itemize}

\textbf{判 0 的情形：}
\begin{itemize}[leftmargin=*,nosep]
    \item 为了强调重点做的有限重复。
    \item 枚举或分步结构里必要的局部重复，但各项信息有实质差异。
    \item 一次性的有效自我纠错、首尾呼应式的一次复述等自然承接，没有造成明显的冗余堆叠。
\end{itemize}

\textbf{示例（判 1）：}问某画的名字。回答先在标题里给出画名和作者，紧接着又用基本信息列表把同样的画名、作者、年代再列一遍，没有新增信息。

\textbf{示例（判 0）：}问某题答案。回答给出推导后，结尾用一句话重申最终答案作为收尾，只重复了一次。

\medskip
\textbf{3. 逻辑性冲突}

\textbf{定义：}模型回答在最终交付内容里，针对同一对象的同一属性、同一个量的取值、或同一个结论，给出两个无法同时为真的说法，且两者都被当作有效信息并列保留，没有择一、没有声明其中一个作废、也没有用条件区分。这是硬矛盾，判 1 时必须能成对定位互斥的两处具体文字 A 和 B。

\textbf{构成硬矛盾要同时满足：}A、B 针对同一对象或同一量或同一结论；处在同一条件、同一时间、同一口径下；逻辑上无法同时为真；模型没有说明其中一个已作废或只是备选；A、B 都在回答里作为有效信息并列保留。

\textbf{判 1 的四类情形（要能指出 A、B 两处具体文字）：}
\begin{enumerate}[leftmargin=*,nosep]
    \item 同一最终结论并列互斥：对同一问题、同一对象或同一答案位置给出两个无法同时成立的最终结论，且没有择一。例如问藏品名称，先说是青铜尊，最后又写最终答案青铜鼎；又如先说 A 正确、B 不符合条件，最终答案却选 B。
    \item 同一对象属性互斥：对同一对象的同一属性给出无法同时成立的判断，且两处都保留使用。例如华容道里先说 b 是横向木块，移动步骤里又写移动竖向木块 b；又如先说某函数是奇函数关于原点对称，又说它关于 y 轴对称所以是偶函数。
    \item 同一量取值互斥：对同一个数值、日期、坐标、数量、角度、面积、温度给出两个无法同时成立的取值，且没有说明修正关系。例如同一底面积先写成一个值后又写成另一个值且都在用；又如说图中共 6 个圆，分项却给出左 3 右 4 共 7 个，总数口径互斥；又如点 P 坐标先写成 (2, 3) 后又写成 (3, 2)。
    \item 推导结论与最终答案互斥且推导未被放弃：推导中明确得到并保留 X，最终答案却给出与 X 互斥的 Y，且没有声明前文作废。例如步骤里说 8、9 是必经点，最终必经点列表却不含 8、9；又如推导得 x 等于 3 并代入验证成立，最终答案却写 x 等于 5。
\end{enumerate}

\textbf{判 0 的情形：}
\begin{itemize}[leftmargin=*,nosep]
    \item 已修正、已放弃或已择一：先出现一种说法，随后明确否定、修正、放弃，或在多个候选中明确选定，最终只保留唯一结论（这类过程摇摆归思维链外溢）。
    \item 模糊、推测、对冲表达：用可能、大概、更像、倾向于、不排除等表示候选或置信度差异，没有形成两个确定互斥的结论。例如说可能是商代晚期，也不排除西周早期。
    \item 条件、时间、视角、口径不同且已区分：例如按假设一结果为 X、按假设二结果为 Y；又如从观察者视角看在左侧、按物体自身朝向看在右侧。
    \item 答案错误、看图错误、计算错误，但内部没有成对互斥的并列说法：本类不判断答案对错，看错图、数错数、算错值，只要内部没有对同一对象或同一量或同一结论并列保留两个互斥说法，就判 0。
    \item 表达不严谨、同义改写、单位换算、重复说明、推理粗糙、伪装推理，但指不出两处并列保留且无法同时成立的具体说法。例如说是高积云即鱼鳞云（分类层级或俗称），又如说 1 米也就是 100 厘米（等价换算）。
\end{itemize}

\textbf{关键区分：}硬矛盾是最终交付内容里对同一对象属性、同一量取值或同一结论并列保留两个无法同时为真的命题，判 1；过程摇摆、模糊措辞、看错算错事实错、条件区分或已择一，都判 0。

\textbf{示例（判 1）：}问从左到右数白头发戴眼镜的人是第几个。回答列表里写第 4 项是目标人物，结尾却写所以是第 5 个，对同一答案并列保留 4 和 5 两个互斥结论且没有取舍。

\textbf{示例（判 0）：}问图中是什么动物。回答从头到尾只说是金毛犬，并补充长毛、金黄色等依据，没有互相否定的说法。

\medskip
\textbf{4. 伪装推理}

\textbf{定义：}模型表面上呈现出分析、推导、论证或证据支撑的样子，但这些内容没有为最终结论提供有效信息增量，只是在制造结果有理有据的表象。

\textbf{有效信息增量}指能支持结论的观察、比较、计数、排除、消歧、计算或核验。在图片任务中，有效增量必须落在图中明确可见的内容上，例如：指出具体物体、文字、颜色、数量、位置；比较多个对象的大小、距离、遮挡、方向、形状、状态；依据图中可见路径、墙体、节点、箭头、表格行列、流程分支判断；对多图人物或物体做去重、匹配、排除；读取图中文字、图例、坐标、标签、数值并用于支撑结论。如果只是套用从图中可以看出、结合整体布局、根据视觉特征、综合判断等话术，却没有指出任何具体可见内容，通常属于伪装推理。

\textbf{判 1 的两类情形（仅当回答已经摆出分析、推导、论证、证据支撑的姿态时）：}
\begin{enumerate}[leftmargin=*,nosep]
    \item 一是无依据的推理包装：表面用了推理或证据支撑的表达，却没有提供任何能支持结论的实质依据。典型表现：用首先、其次、因此、综上、可见等连接词但中间只是空泛表态；只是重复题干、改写结论、复述用户已给信息；图片任务里声称根据图中某某、从视觉上判断、结合空间布局，却不指出具体元素、位置、数量、文字、颜色、方向、遮挡或关系；堆砌从语义结构上下文三方面看等分析框架却不连接到题面或图中具体信息。例如问图中植物是什么，回答说结合叶片形态、生长习性和整体视觉特征可判断是某植物，却不指出任何具体可见特征。
    \item 二是依据不支撑结论：给出一些看似相关的分析或观察，但推不出最终结论，或只是在包装一个本可直接读取、直接观察或用户已给出的答案。典型表现：分析的是 A 结论却落到 B；前文只是背景介绍推不出答案；识别题靠看起来像、风格符合、造型接近等泛泛印象，而不是能确认身份或对象的真实依据；低推理负担任务里答案本可直接观察读取定位，额外分析没有新增观察、比较、计数、排除、消歧或核验；路线规划只复述用户已给顺序，没有比较位置、道路、距离、可达性或障碍；迷宫或路径题只说某点是关键节点、必经节点，却没有用墙体、通道连通性或替代路线排除来证明。例如认人题只凭穿着正式、气质沉稳像某类演员就下身份结论；又如计数题用小狗有四条腿毛茸茸来包装共 4 只这个数。这里还包括识别鉴定类的背景包装：给出名称或身份后主要堆砌生平、作品列表、风格通论、常演角色、历史介绍等与如何从图中确认无关的背景，却不指出图中能锁定结论的可辨识特征（面部、铭文、款识、标志性细节）。
\end{enumerate}

\textbf{判 0 的情形：}
\begin{itemize}[leftmargin=*,nosep]
    \item 直接作答或只做简短说明，没有展开分析来包装结论。注意只说结合图片而不指出图中具体要素，按直接作答处理。
    \item 图片分析有有效可见依据：指出了图中具体可见内容并用来支撑结论，即使简短、概括、\raisebox{-0.08em}{\includegraphics[height=0.95em]{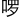}}嗦、不完美。例如说右下角按钮写着提交所以是提交按钮；又如比较两杯水位和容器形状判断哪杯更多。注意：这里要求的是能唯一锁定结论的具体可辨识细节，例如读出文字、铭文、款识，给出确切的数量、位置、可比较的状态；如果只给造型接近、风格符合、科技感、火焰核心、画风色彩相近、看着像这类泛泛特征词，不算有效依据，应按判 1 处理。
    \item 据图做了真实连通性推进但看错（看图错误，判 0）：仅当模型确实逐格交代了墙体或通道是否连通、相邻格能否通行，并据此排除替代路线（即给出了能支撑路径结论的连通性核验），只是把某段墙体或通道看错时，才算看图错误判 0。如果只是泛泛点名某些格子、节点为关键点或必经点、或只复述各数字所在位置，而没有用连通性或排除来证明，仍按上面的依据不支撑结论判 1。
    \item 用户明确要求说明理由、分析过程、逐步判断或给出依据，且分析不是完全空转。例如用排除法说明为什么选 C。
    \item 看错、算错、事实错，但依据和结论之间存在真实支撑关系。例如基于图中真实对象计数，只是漏数或数错。
    \item 看不清图片时的谨慎说明：明确表示图片模糊、无法确认，并给出谨慎结论或拒绝断言。
\end{itemize}

\textbf{关键区分：}直接答题、有效图像依据、有效推理、思维链外溢都判 0；看似在分析实则重复题干、复述结论、堆连接词、泛泛表态，或声称结合图中具体特征却不给任何具体可见细节，判 1；依据可能错但确实在尝试基于图片或题面支撑结论的，判 0。不要因为回答长、步骤多、语气像推理就判 1，也不要因为结论错就判 1。

\textbf{示例（判 1）：}问图中哪个按钮是提交按钮。回答说需要结合界面布局、按钮语义和操作路径综合判断，因此是右下角那个，没有指出按钮文字、颜色或图标等具体可见依据。

\textbf{示例（判 0）：}问图中有几个苹果。回答说左侧 1 个、右侧盘子里 2 个、一共 3 个，给出了具体的图像观察和计数依据。

\medskip
\textbf{四、输出要求}

请严格按下面顺序和格式输出，不要输出无关内容，不要省略任何类别。每一类先给 1 或 0 的判定，再用一句话说明理由（判 1 时点出关键依据，逻辑性冲突要点出互斥的 A、B 两处）。

\medskip
{\ttfamily
1. 思维链外溢：1 / 0\\
理由：\\[2pt]
2. 病态重复：1 / 0\\
理由：\\[2pt]
3. 逻辑性冲突：1 / 0\\
理由：\\[2pt]
4. 伪装推理：1 / 0\\
理由：
}

\medskip
\textbf{五、输入}

\textbf{问题：}\\
\texttt{\{question\}}

\medskip
\textbf{模型回答：}\\
\texttt{\{response\}}
\end{CJK*}
\end{AIbox}

\end{document}